\pdfoutput=1

\documentclass[11pt]{article}

\usepackage{ACL2023}

\usepackage{times}
\usepackage{latexsym}
\usepackage{graphicx}
\usepackage{xurl}
\newcommand{\bs}{\textcolor{black}}

\newcommand{\red}{\textcolor{red}}

\newcommand{\green}{\textcolor{green}}

\usepackage{epigraph}
\usepackage{graphicx}
\usepackage{caption}
\usepackage{subcaption}

 \usepackage{booktabs}

\definecolor{ywl}{HTML}{FFF2CC}
\definecolor{prl}{HTML}{E1D5E7}
\definecolor{grn}{HTML}{D5E8D4}
\definecolor{grk}{HTML}{E6E6E6}
\definecolor{fem}{RGB}{32, 178, 170}
\definecolor{masc}{RGB}{200, 162, 200}

\usepackage{tabularx}
\usepackage{enumitem}
\usepackage{float}
\usepackage{amssymb}

\usepackage{tcolorbox}
\usepackage{colortbl}
\usepackage{amsmath}
\usepackage{multirow}

\definecolor{mygray}{RGB}{220,220,220}
\definecolor{mydarkgray}{RGB}{240,240,240}

\usepackage{twemojis}
\newcommand{\finding}{\scalebox{1.5}{\twemoji{1f50d}}}

\usepackage[T1]{fontenc}

\usepackage[utf8]{inputenc}

\usepackage{microtype}

\usepackage{inconsolata}

\title{What the Harm? Quantifying the Tangible Impact of Gender Bias in Machine Translation with a Human-centered Study}

\author{Beatrice Savoldi\textsuperscript{\textcolor{fem}{$\blacksquare$}}, Sara Papi\textsuperscript{\textcolor{fem}{$\blacksquare$}}, Matteo Negri\textsuperscript{\textcolor{fem}{$\blacksquare$}},\\ \textbf{Ana Guerberof Arenas\textsuperscript{\textcolor{masc}{$\bigstar$}}, Luisa Bentivogli\textsuperscript{\textcolor{fem}{$\blacksquare$}}} \\
  \textsuperscript{\textcolor{fem}{$\blacksquare$}}Fondazione Bruno Kessler, Italy \\
  \textsuperscript{\textcolor{masc}{$\bigstar$}}University of Groningen, Netherlands \\
  \texttt{\{bsavoldi,spapi,negri,bentivo\}@fbk.eu} \\
  \texttt{a.guerberof.arenas@rug.nl}}

\begin{document}
\maketitle
\begin{abstract}

Gender bias in machine translation (MT) is recognized as an issue that can harm people and society. And yet, advancements in the field 
rarely involve people,
the final MT users,
or inform how they might be impacted by biased technologies. Current evaluations are often restricted to 
automatic methods, which 
offer an opaque
estimate of what the 
downstream 
impact of gender disparities might be. 
We conduct an 
extensive 
human-centered 
study to examine if and to what extent bias in MT 
brings harms with tangible  costs, 
such as quality of service gaps across women and men. To this aim, we collect behavioral data from  $\sim$90
participants, who post-edited MT outputs to ensure correct gender translation. 
Across multiple datasets, languages, and types of users, our study  
shows that feminine post-editing demands significantly more technical and temporal effort, also corresponding to higher financial costs. 
Existing bias measurements, 
however, fail to reflect the found disparities. 
Our findings advocate for human-centered approaches that can inform the societal impact of bias. 


%

\end{abstract}

\section{Introduction}
\label{sec:intro}


\begin{figure}[htp]
    \centering
    \includegraphics[width=\linewidth]{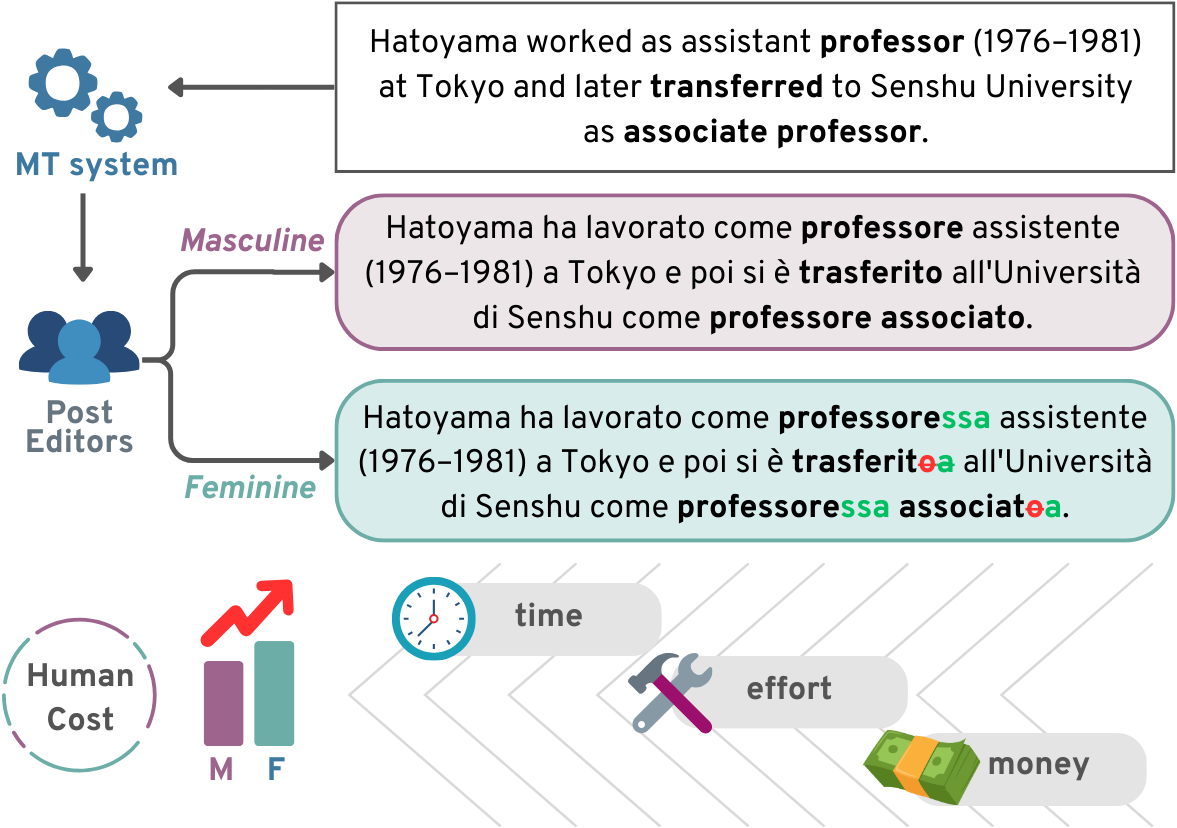}
    \caption{Harms as assessed in our study design. We task participants
    with the post-editing of an MT output into both feminine and masculine gender. We collect behavioural data (i.e. time and technical effort) and assess 
    higher workload and economic costs 
    associated with feminine translations.}
    \label{fig:overview}
\end{figure}

Natural language processing (NLP) has evolved from an academic specialty to countless commercial applications that can both benefit and negatively affect people's lives. With the widespread use of these technologies, researching the ethical and social impact of NLP has become increasingly crucial \citep{hovy-spruit-2016-social, sheng-etal-2021-societal}, with gender fairness being a major concern \citep{sun-etal-2019-mitigating, stanczak2021survey}.

In machine translation (MT)
gender bias has received significant attention, 
also in the public domain \citep{forbes}. Numerous studies have shown that MT perpetuates harmful stereotypes \citep{stanovsky-etal-2019-evaluating, triboulet-bouillon-2023-evaluating} and is skewed towards 
masculine forms that under-represent women \citep{vanmassenhove-etal-2018-getting,alhafni-etal-2022-user}. 

As emphasized by \citet{savoldi-etal-2021-gender} – if we regard MT as 
a 
resource 
in its own right -- such representational disparities might directly imply \textit{allocative} harms, i.e. differential access to material benefits that make a social group or individual worse-off \citep{barocas,chien2024beyond}.
For instance, a woman using an MT system to translate her biography (i.e. the first sentence in English in Figure \ref{fig:overview}) into Italian would need more effort (i.e. represented by 
insertions
-- in green, and substitutions -- in red and green -- in Figure \ref{fig:overview}) to revise incorrect masculine references, thus 
experiencing a 
disparity in the 
quality of the service. 

Despite acknowledging the potential harm to individuals,
research on gender bias in MT primarily focuses on in-lab automatic evaluations. Such assessments, however, are only assumed to reflect a real-world downstream effect, without verifying if and to what extent biased models might concretely impact users interacting with a system. 
%


To address this gap, we examine the effect of gender bias in MT with a human-centered perspective. 
Specifically, 
we ask: \textbf{\textit{Does gender bias in MT imply tangible service disparities across men and women?}} And if so, can we meaningfully quantify them via more human-centered measures?
%
To take stock of the current research landscape, 
we 
review 
the involvement of human subjects in prior literature on gender 
and MT. Motivated by the outcome, we conduct extensive experiments across multiple datasets, languages, and 
users. 
%
In a controlled setup, 88 participants post-edited 
MT outputs
to ensure either feminine or masculine 
gender 
translation.\footnote{We discuss the implications this binary  setup in \S\ref{sec:ethical}.} 
In the process, we track behavioral data -- i.e. time to edit and number of edits -- to compare effort across genders. Based on this, we estimate the associated cost for 
post-editing into each gender
if the work were 
assigned to a third-party translator. Our main findings are:
\smallskip

\begin{enumerate}[itemsep=0.5ex, parsep=0.5pt, partopsep=0.5pt, topsep=0.5ex] 
\item Most of current assessments of gender bias in MT either overlook 
human involvement, or treat individuals as models' evaluators rather than potentially affected users (\S\ref{sec:review}).


\item We find substantial gender disparities in the time and technical 
effort required to post-edit MT, with feminine translation taking on average twice longer and four times the editing operations  
(\S\ref{sec:results}).
\item The cost of the found disparities is also economic, and can unfairly fall onto various stakeholders 
in the translation process  (\S\ref{sec:discussion}).


\item  The automatic evaluation of gender bias does not accurately reflect the found human-centered  effort disparities (\S\ref{sec:discussion}). 

\end{enumerate}




To sum up, our
study marks a step towards understanding the implications of gender bias in MT. While harms have so far been implied, or inferred from automatic scores as a proxy for downstream impact, here we empirically show that gender bias can bring unfair service disparities. What's more, we quantify bias with measures that are more  meaningful for potentially impacted individuals: workload and economic costs. 

\bs{Behvaioural data and post-edits are made available at \url{https://huggingface.co/datasets/FBK-MT/gender-bias-PE}.}


\section{Where are the people? \\ \texttt{Evaluator != User} }
\label{sec:review}

Language technologies have reached a level of quality that enabled laypeople to integrate them into their day-to-day activities \citep{nurminen-papula-2018-gist}. With this shift, understanding users' needs, and how they might be impacted becomes of utmost importance. Indeed, NLP is witnessing
increasing 
emphasis 
towards more human-centered approaches\footnote{A case in point being the introduction of the ``human-centered NLP'' track in ACL* conferences.}
\citep{robertson2021three, goyal2023else}, but still little is known about the experience of 
people interacting with such technology – even for wide-reaching, user-facing applications
such as MT \citep{guerberof2023ethics}.\footnote{\citet{briva2023impact} claim that also for 
professional translators existing studies mostly focus on industry-oriented productivity gains rather than on user experience.}

Similarly, the study of bias is 
emphasized as an intrinsically human-centered endeavour  \citep{bendertypology}
that requires understanding which behaviour might be harmful, how and to whom \citep{Blodgett-etal-2020-language}. Nonetheless, there is a paucity of work that foregrounds human engagement 
\citep{cercas-curry-etal-2020-conversational, mengesha2021don, wang2024measuring}. 
Arguably, truly informative measurements on the downstream effects of bias and its potential for harm should assume  people as 
target.
%
%
%
%
%
But in what capacity, if any, have people  been involved so far in the study of gender bias in MT?




\paragraph{ACL Anthology search}
For a systematic review of prior work, we query the ACL anthology.\footnote{\url{https://aclanthology.org/}}
As keywords, 
we specify our application of interest -- e.g. ``MT'' and ``translation'' -- 
combined
with 
``gender'' 
or ``bias''. For a more channelled query focusing on 
people involved in bias assessment, we also add other human-engagement dedicated keywords (e.g. ``user'', ``survey''). As of April 2024, our search returned 251 articles, 96 of which also matched the human-engagement keywords.
Upon manual inspection,
we retained \bs{105}
\textit{in-scope} manuscripts,\footnote{Works focusing on (human) gender translation, bias or fairness in the context of MT.} 
and discarded unrelated papers  focusing on other 
definitions
of the keywords (e.g. ``inductive bias''). 
The in-scope papers were finally reviewed by focusing on the 
presence,
or lack thereof, of human involvement. 
For more details on our search and selection, see 
Appendix \ref{app:acl_query}.  

\begin{figure}[t]
  \centering
\includegraphics[scale= 0.43]{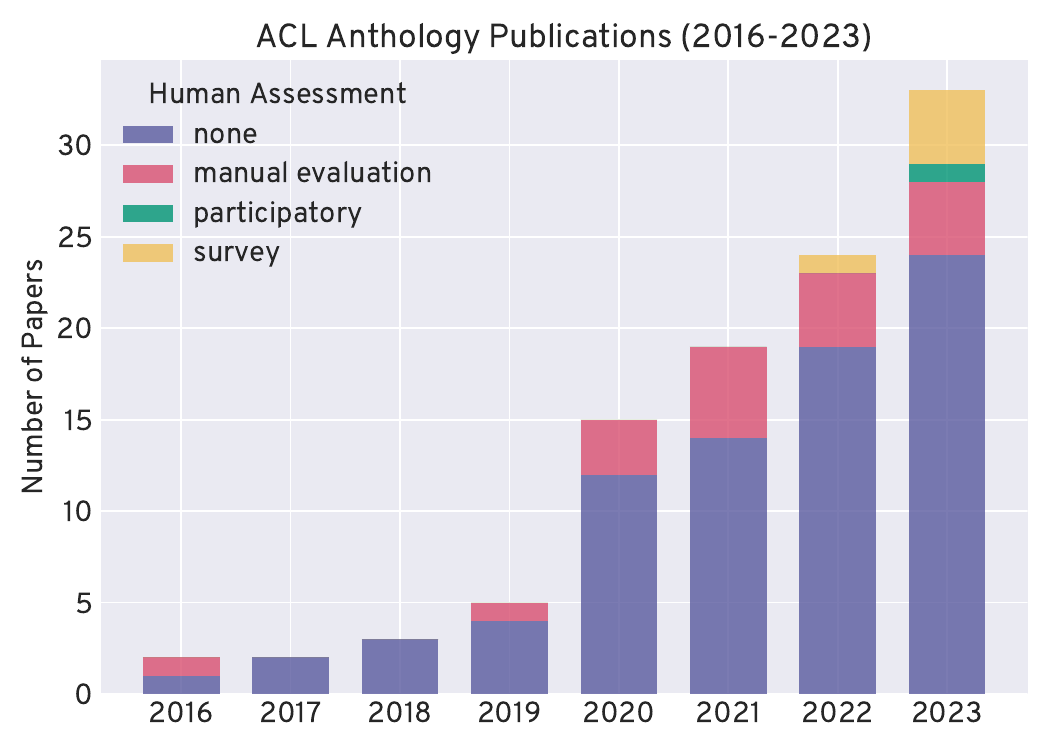}
  \caption{Human involvement in the assessment and framing of gender (bias) in MT, based on an ACL Anthology search. For 
  studies with human participants,
  we distinguish
  qualitative, but yet model-centric \textsc{manual evaluation}, and more human-centric designs -- i.e. \textsc{survey} studies and \textsc{participatory} approaches.}
  \label{fig:human-acl}
\end{figure}

\paragraph{Review} We report the results of our review 
in Figure \ref{fig:human-acl}. As the image shows, we attest to a steady increase 
in publications related to gender in MT, in particular from 2020 onwards. In line with our expectations and the general trend in NLP, however, \textbf{we attest a severe lack of human engagement}. 
In fact, only \bs{24} works rely on humans to measure bias, though in a different capacity, which we distinguish into three conceptual categories. In \bs{18} papers, we find that people -- 
often
expert linguists (e.g. \citet{vanmassenhove-etal-2021-neutral, soler-uguet-etal-2023-enhancing}) -- are involved 
in
\textsc{manual evaluation}. This serves to either ensure correlation with bias metrics (e.g. \citet{kocmi-etal-2020-gender}) or to gain qualitative insights that defy automatic approaches \citep{popovic-2021-agree}. 
While indeed valuable, such analyses  
are 
a support for structured, often annotation-based \textit{model-centric} evaluations -- i.e. 
that inform and quantify models' behaviour. 
Differently, the \bs{5} papers in the \textsc{survey} category focus on the feedback and experiences of potentially impacted groups of users (e.g. \citet{piergentili-etal-2023-hi, daems-hackenbuchner-2022-debiasbyus}).
For instance, to grasp 
user preference in how  
models
should handle the translation of novel, non-binary pronouns from English -- e.g. \textit{ze, xe} \citep{lauscher-etal-2023-em}\footnote{Interestingly, all \textsc{survey} works focus on non-binary linguistic strategies beside feminine/masculine ones. See \S\ref{sec:ethical}.},  or to understand the potential trade-off between overall quality and inclusivity goals \citep{amrhein-etal-2023-exploiting}.
Finally, the study by \citet{gromann-etal-2023-participatory} recounts a \textsc{participatory} Action Research, where a community-led approach with different stakeholders informs the state and potential direction for gender fair MT. 


Overall, despite this recent trend towards surveys or participatory methods, humans are rarely involved to estimate gender bias in MT. 
Moreover, \textbf{if involved,}
\textbf{people mostly serve in the capacity of \textit{evaluators}}, supporting  model-centric assessments \textbf{rather than being considered as potentially impacted \textit{users}}. Our finding stands in contrast with a qualitative survey by \citet{dev-etal-2021-harms}, which found MT as \bs{an application with a high risk for downstream harms in the context of gender bias.} 

Further motivated by such evidence, we carry out a quantitative, empirical study -- to the best of our knowledge, the very first of its kind -- focusing on human-centered assessments. In particular, 
we examine whether gender bias in machine translation leads to disparities in the quality of service offered to women and men, by considering different \textit{datasets, languages}, and \textit{users} (\S\ref{subsec:settings}).



\section{Experimental setup}
\label{sec:experimental}

We simulate the conditions in which 
an end user needs
the translation of a text referring to them -- as described in \S\ref{sec:intro} and exemplified in Figure \ref{fig:overview}.  
To strike a  balance between 
controlled conditions for reliable findings while  
keeping  
a realistic scenario, the study is realized as a \textit{post-editing task} (PE), where participants are asked to 
also ensure
that human references are rendered as either feminine or masculine. The same output sentences are edited twice (once per gender), thus allowing to isolate any difference in effort as a gender-related factor. 

\bs{Note that our experiments are based on sentences that always require to translate gender and enable focused analyses. As we further discuss in \ref{sec:limitations}, we thus mimic scenarios that often require to manage gender mentions to human referents, as in the case of biographies, CVs, and administrative texts.}

\subsection{Settings}
\label{subsec:settings}

\paragraph{Languages}
We include three language pairs -- English$\rightarrow$Italian/Spanish/German -- which are representative of the challenges of translating into languages with extensive gendered morphology -- e.g. \textit{the friend}$\rightarrow$ es: \textit{\textbf{el}/\textbf{la}  amig\textbf{o}/\textbf{a}}. Overall, these pairs feature sufficiently diverse gender phenomena \citep{gygax}. 
The selection was also bound to their representation within available datasets.   

\paragraph{Datasets}
We rely on MT datasets that represent naturally occurring gender translation phenomena. Namely, MT-GenEval \citep{currey-etal-2022-mt} -- which is built upon Wikipedia biographies -- and the TED-derived Must-SHE corpus \citep{bentivogli-etal-2020-gender}. Our data samples are organized as follows. \textbf{\textit{(i)}} \textsc{mtgen-a}, a subset of MT-GenEval sentences where gender in the source is ambiguous.\footnote{e.g. ``\textit{Hatoyama worked as assistant professor}[...]''} \textbf{\textit{(ii)}} \textsc{mtgen-un}, which contains feminine/masculine versions of gender-unambiguous English sentences,\footnote{e.g. ``\textit{\underline{She} was appointed Archdeacon of Lismore} [...]'' vs. ``\textit{\underline{He} was appointed Archdeacon of Lismore}[...].''} thus offering favourable conditions for correct 
translation based on available gender cues in the source. 
Finally, \textbf{\textit{(iii)}} a subset of \textsc{must-she} featuring ambiguous first-person references in the English source sentence.\footnote{\textit{``I immediately began to doubt myself} [...]''} 
This sample is included for the sake of 
phenomena variability -- given that \textsc{must-she} entails gendered translation for many parts-of-speech  --whereas both Wikipedia-derived samples mainly represent gendered translations for occupational nouns. 

As a key feature of these datasets, for each English source sentence, two contrastive feminine/masculine pairs of reference translations are provided. These are designed to isolate gender as a factor from overall quality evaluation.\footnote{These references allow us to compare human-centered results with those of automatic metrics in \S\ref{sec:discussion}. We adjusted a few inconsistencies in \textsc{mtgen-a} references -- see \ref{app:ref}.}

As 
described in \S\ref{sec:results}, 
we conduct 
multi-\textit{dataset} (\S\ref{subsec:multidata})
experiments for en-it, whereas the 
multi-\textit{language} (\S\ref{subsec:multilang})
study with en-es/de is based on 
\textsc{mtgen-a}. 
For each dataset (statistics in Table \ref{tab:data_stats}),
we retrieve a 
random sample of 250 sentences, while maximizing the number of common sentences across language pairs.\footnote{See Appendix \ref{app:data_info} for details on sample extraction.} 


\begin{table}[t]
\centering
\footnotesize
\begin{tabular}{llccc}
\toprule
\textbf{} & & \# src-W & \# out-W & \# tgt-GW \\
\midrule
\textit{en-it} & \textsc{mtgen-un}                       & 24 & 25 & 4.57 \\
\textit{en-it} & \textsc{must-she}                            & 25 & 24 & 1.58 \\
\textit{en-it} & \textsc{mtgen-a} & 17 & 17 & 2.38 \\
\textit{en-es} & \textsc{mtgen-a} & 18 & 19 & 2.34 \\
\textit{en-de} & \textsc{mtgen-a} & 17 & 17 & 2.61 \\
\bottomrule
\end{tabular}
\caption{Data statistics. For each dataset and language, we provide the average number of words for source \textbf{(src-W)} and output sentences \textbf{(out-W)}, as well as the average number of target gendered words \textbf{(tgt-GW)} in the reference translations.}
\label{tab:data_stats}
\end{table}

 \paragraph{User types}
%
The study aims to reflect an average user, 
who fixes an MT output by themselves. While including lay users with different levels of language expertise or MT familiarity would represent a comprehensive case study, such a setup adds a notable level of complexity and potential noise to deal with (e.g. gendered expressions to be fixed might be overlooked). To guarantee higher control of our variables, we thus rely on professional translators as a proxy. Still, to also mimic MT interactions with less experienced users, for en-it we carry out multiuser experiments (\S\ref{subsec:multiuser}) involving high-school students, 
native speakers of Italian with a B2 level of English 
(further details in 
the upcoming 
\S\ref{subsec:study_design}). 
To avoid fitting our results to the potentially subjective post-editing activity of one person, we allocate 16 post-editors for
\textsc{must-she} and 16 for \textsc{mtgen-un}. 
Since it consists of shorter sentences (see Table \ref{tab:data_stats}), we task 14 subjects for each of the  four \textsc{mtgen-a} conditions  -- for a total of 88 participants overall. 


\paragraph{Model}
Reliable behavioural assessments require a sizable data sample and number of participants, which we prioritize during budget allocation.
For this reason, we do not consider MT models as a variable and only use Google Translate (GT).
Besides being  
state-of-the-art,
GT is chosen as it represents one of the most widely used consumer-facing commercial MT systems \citep{pitman2021translate}.


\begin{table*}[htp]
\addtolength{\tabcolsep}{-1.5pt}
\footnotesize
\centering
\begin{tabular}{lll|rrrr||rrrr||rrrr}
\toprule
& & &\multicolumn{4}{c||}{\textbf{\textsc{TE ($ \downarrow $)}}}                           & \multicolumn{4}{c}{\textbf{\textsc{HTER ($ \downarrow $)}}} 
 &  \multicolumn{4}{c}{\textbf{\textsc{\# Edited sent ($\downarrow $)}}}                
 \\
 User & Lang & Dataset & \textsc{Fem} & \textsc{Mas} & $\Delta$\textit{abs} & $\Delta$\textit{rel}  & \textsc{Fem} & \textsc{Mas} & $\Delta$\textit{abs} & $\Delta$\textit{rel}  & \textsc{Fem} & \textsc{Mas} & $\Delta$\textit{abs} & $\Delta$\textit{rel}  
\\
\toprule
\toprule
 P & \textit{en-it} & \textsc{mtgen-un}   
&  2:58 & 2:11 
& \cellcolor{mygray}0:47  & \cellcolor{mydarkgray}36.3
& 8.17  & 5.16 & \cellcolor{mygray}3.01  & \cellcolor{mydarkgray}58.3 
& 142 & 83 & \cellcolor{mygray}59 &  \cellcolor{mydarkgray}71
\\

 P & \textit{en-it} & \textsc{must-she}           
&  2:33 & 1:27 
&\cellcolor{mygray}1:06 & \cellcolor{mydarkgray}76.1
& 8.07 & 3.16 & \cellcolor{mygray}4.91  & \cellcolor{mydarkgray}155.4 
& 226 & 58 &\cellcolor{mygray}168 &\cellcolor{mydarkgray}290
\\
 P & \textit{en-it} & \textsc{mtgen-a}            
&  2:38 & 0:57  
& \cellcolor{mygray}1:41 &  \cellcolor{mydarkgray}177.6
& 16.51 & 5.47 & \cellcolor{mygray}13.08 & \cellcolor{mydarkgray}201.8  
& 243  & 70 &\cellcolor{mygray}173 & \cellcolor{mydarkgray}247
\\
\midrule
 P & \textit{en-es} & \textsc{mtgen-a}            
& 2:13  &  1:13
& \cellcolor{mygray}0:59 &  \cellcolor{mydarkgray}81.1
& 14.88  & 5.76  & \cellcolor{mygray}9.12 & \cellcolor{mydarkgray}158.3  
& 242 & 93 & \cellcolor{mygray}149 & \cellcolor{mydarkgray}160
\\

 P & \textit{en-de} & \textsc{mtgen-a}            
&  2:12 & 0:30
& \cellcolor{mygray}1:42 &  \cellcolor{mydarkgray}334.0
& 15.62 &  5.47 & \cellcolor{mygray}11.04 & \cellcolor{mydarkgray}515.0  
& 228 & 40 & \cellcolor{mygray}188 & \cellcolor{mydarkgray}470
\\ 

\midrule

S & \textit{en-it} & \textsc{mtgen-a}            
& 2:08  &  0:29
& \cellcolor{mygray}1:38 &  \cellcolor{mydarkgray}329.8
& 13.18 & 1.79 & \cellcolor{mygray}11.39 & \cellcolor{mydarkgray}636.3  
& 242 & 38 & \cellcolor{mygray}204 & \cellcolor{mydarkgray}573
\\
\midrule
\midrule

& & \textbf{AVG.} & 2:27 & 1:08 & \cellcolor{mygray}1:19 & \cellcolor{mydarkgray}116.2
& 12.74 & 3.98 & \cellcolor{mygray}8.76 & \cellcolor{mydarkgray}220.1 
& 221
  & 64 & \cellcolor{mygray}157 & \cellcolor{mydarkgray}245
\\

\bottomrule
\end{tabular}
\caption{Multidataset (top), multilanguage (center) and multiuser (bottom) results. 
\bs{Results are shown for all users -- both (P)rofessional and (S)tudents -- languages, and datasets.}
We provide time to edit (TE, i.e. hour:minutes), HTER, and the number of post-edited sentences (out of 250 per each gender).}
\label{tab:absolute_results}
\end{table*}

\subsection{Study design}
\label{subsec:study_design}

\textbf{Task instructions and platform}
Given a source sentence and its MT output, 
participants were instructed\footnote{For each condition, we prepared dedicated guidelines, which are available at \url{https://github.com/bsavoldi/post-edit_guidelines}}
to carry out a \textit{light} PE -- i.e. targeting only essential fixes to adjust the overall quality of the translation \citep{o2022deal}
-- with a 
focus on ensuring either feminine or masculine translation for human referents, based on provided gender information. We choose a 
light
PE \textit{i)} given the 
high quality of the MT output,\footnote{E.g., COMET scores are between 82.3-85.3 across all languages and data. See Appendix \ref{app:overall-translation-quality} for full results.}
and crucially \textit{ii)} to limit the number of preferential edits that might introduce noise. 
The task was performed with Matecat,\footnote{\url{https://www.matecat.com/}} a mature,  
computer-assisted translation (CAT) tool supporting PE that is freely available online.\footnote{For more details on the Matecat setup see 
Appendix \ref{app:matecat}.}

\paragraph{Within-group design}
For each data sample of 250 <English source, GT output> pairs, we design a within-subjects study with counterbalancing \citep{charness2012experimental}, \bs{which ensures  variation of the order of conditions in the study.}
Namely, each participant performs \textit{i)} both feminine (F) and masculine (M) post-edits, \textit{ii)} in equal amounts (blocks of around 15 sentences each),  \textit{iii)} balancing at the sample level which
block -- F or M -- they will post-edit first. 
A within-subject approach is ideal to distribute potential extraneous effects (e.g. participants' tendency to edit more or take longer) across F and M post-edits. Also, counterbalancing handles carryover effects such as \textit{order} and \textit{fatigue}\footnote{For fatigue, we also only assign 30 sentences per subject.}
\citep{price2017research}. Crucially, to control for \textit{familiarity} effects, we ensure that a participant never post-edits the same output twice across genders. 

The design remains the same for all samples, but 
always involving different subjects, so as to ensure the generalization and replicability of our results. 

\paragraph{Participants recruitment and task organization}
Experiments for en-it include data from 
both \textit{i)} professional translators based on voluntary participation, and \textit{ii)} paid professionals. We attested no significant difference between these conditions (for further details see Appendix \ref{app:agreement}).  
For en-de/es, we exclusively relied on paid professionals. 
Experiments were allocated 50m (i.e.~$\sim$10m 
instructions and $\sim$40m PE), which vastly ensured the 
sufficient time to complete
the task.\footnote{
Based on industry standards, we estimated a PE speed of $\sim$25 words per minute.}
The experiment with students was carried out as part of their school activities:
we allocated double the time and included a warm-up phase to get acquainted with the PE task.
No participant was informed of the scope of our study beforehand, and all recorded data are anonymous. 
For further information on the recruited participants and compensation 
see Appendix \ref{app:participants}.





\begin{figure*}[t]
    \centering
       
    \begin{subfigure}[t]{0.325\textwidth}
        \includegraphics[width=\textwidth]{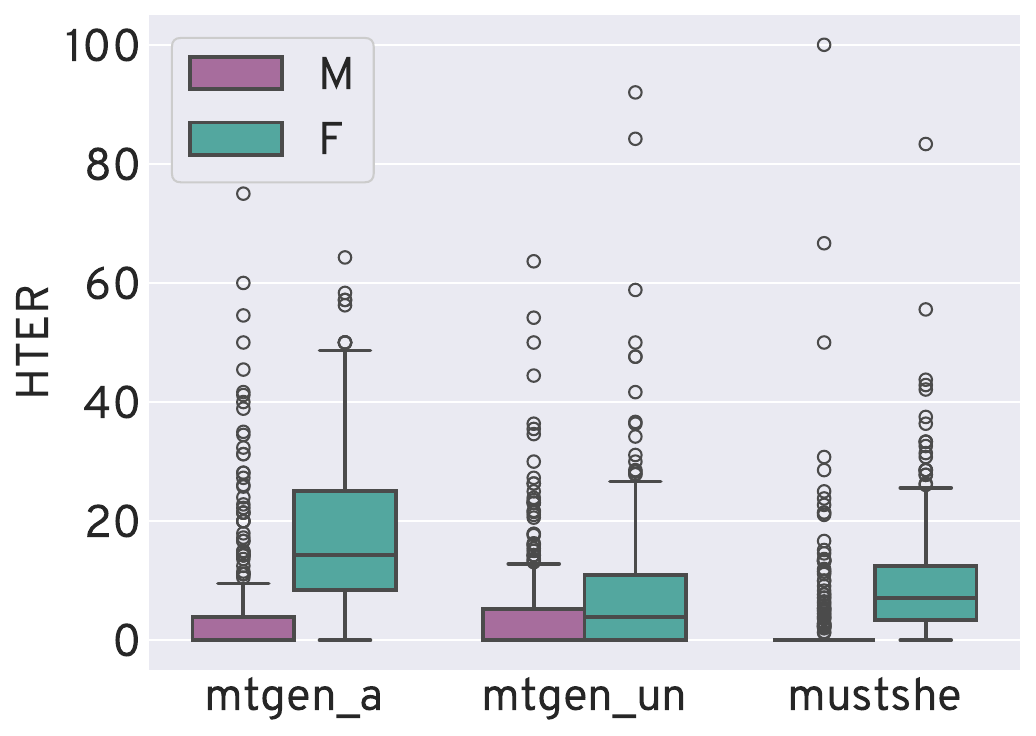}
        \caption{Multidataset}
        \label{fig:hter_multidataset_sent}
    \end{subfigure}  
    \begin{subfigure}[t]{0.325\textwidth}
        \includegraphics[width=\textwidth]{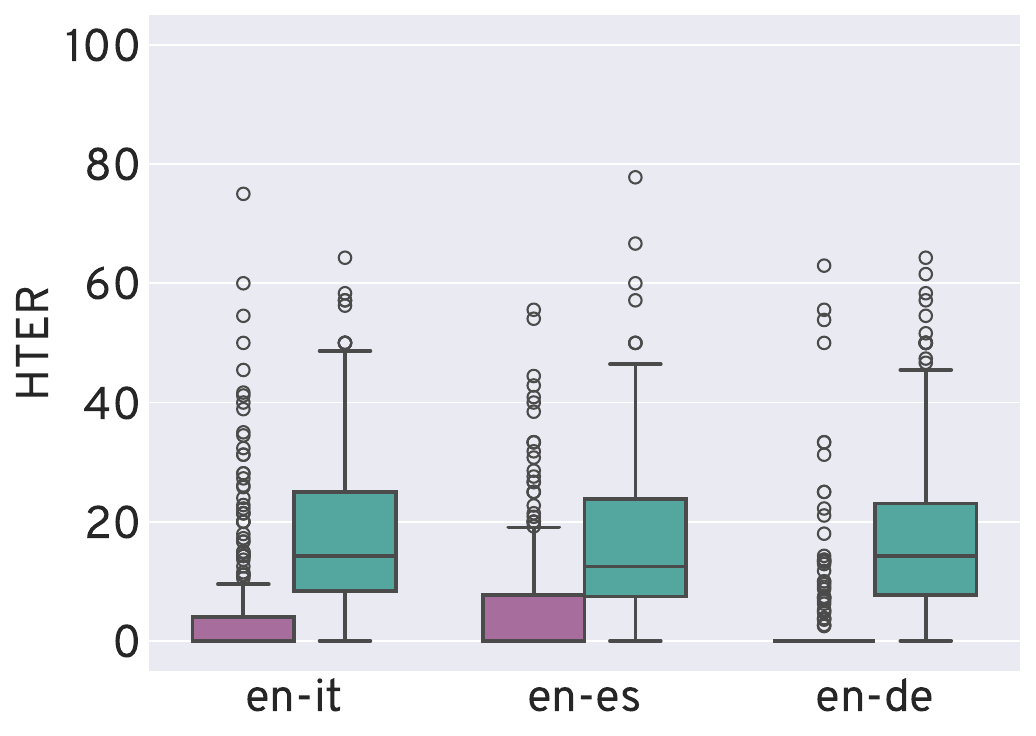}
        \caption{Multilanguage}
        \label{fig:hter_multilang_sent}
    \end{subfigure}   
    \begin{subfigure}[t]{0.325\textwidth}
        \includegraphics[width=\textwidth]{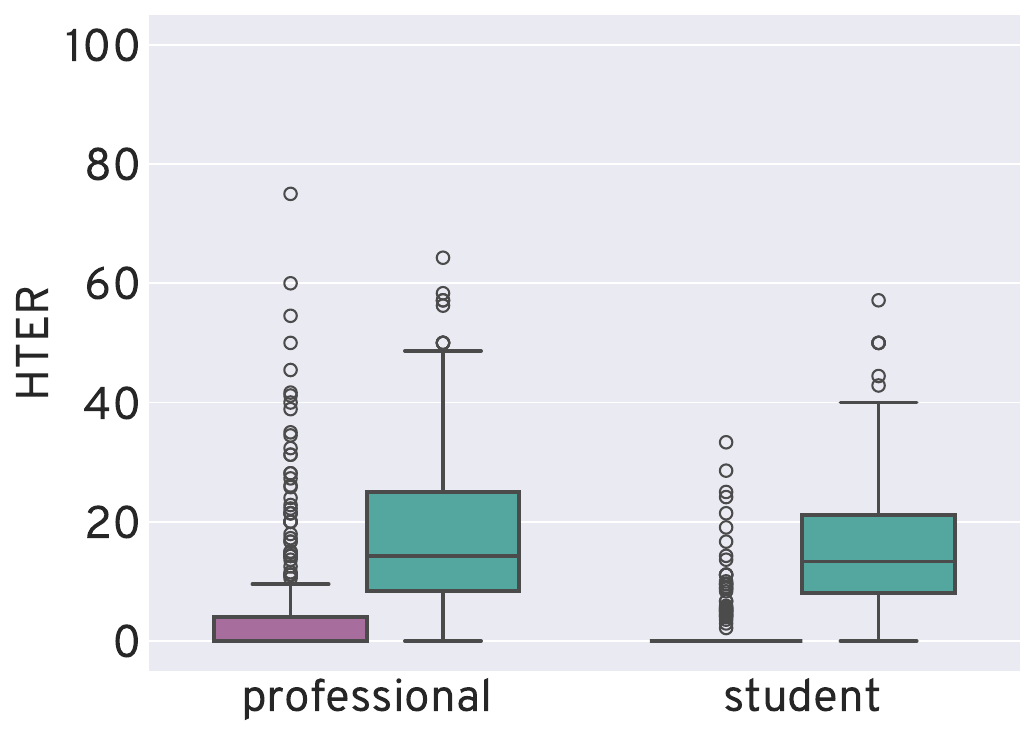}
        \caption{Multiuser}
        \label{fig:hter_multiuser_sent}
           \end{subfigure}
        
    \caption{HTER distribution across post-edited sentences.}
    \label{fig:hter_sent}
\end{figure*}


\begin{figure*}[t]
    \centering
       
    \begin{subfigure}[t]{0.325\textwidth}
        \includegraphics[width=\textwidth]{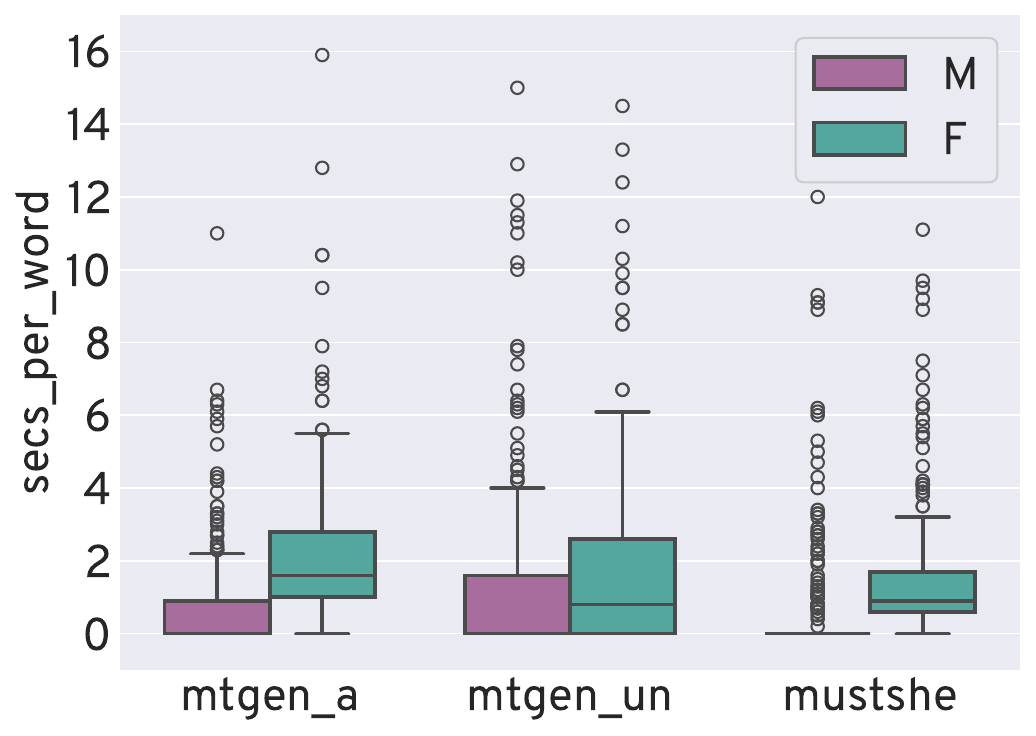}
        \caption{Multidataset}
        \label{fig:sec_multidataset_sent}
    \end{subfigure}  
    \begin{subfigure}[t]{0.325\textwidth}
        \includegraphics[width=\textwidth]{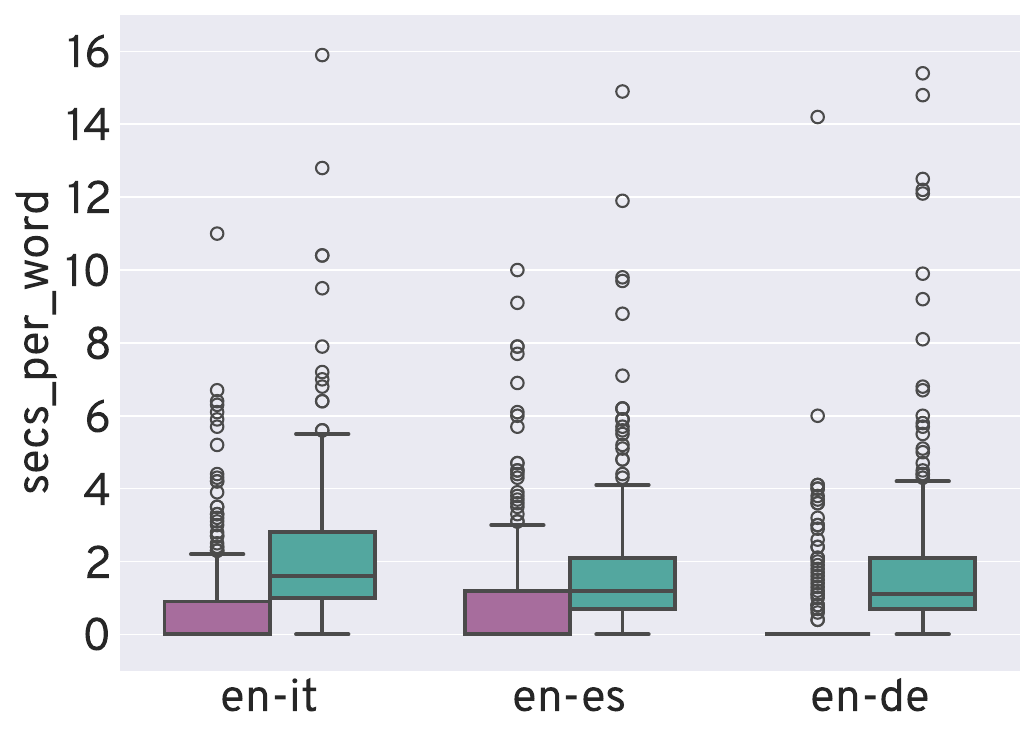}
        \caption{Multilanguage}
        \label{fig:sec_multilang_sent}
    \end{subfigure}   
    \begin{subfigure}[t]{0.325\textwidth}
        \includegraphics[width=\textwidth]{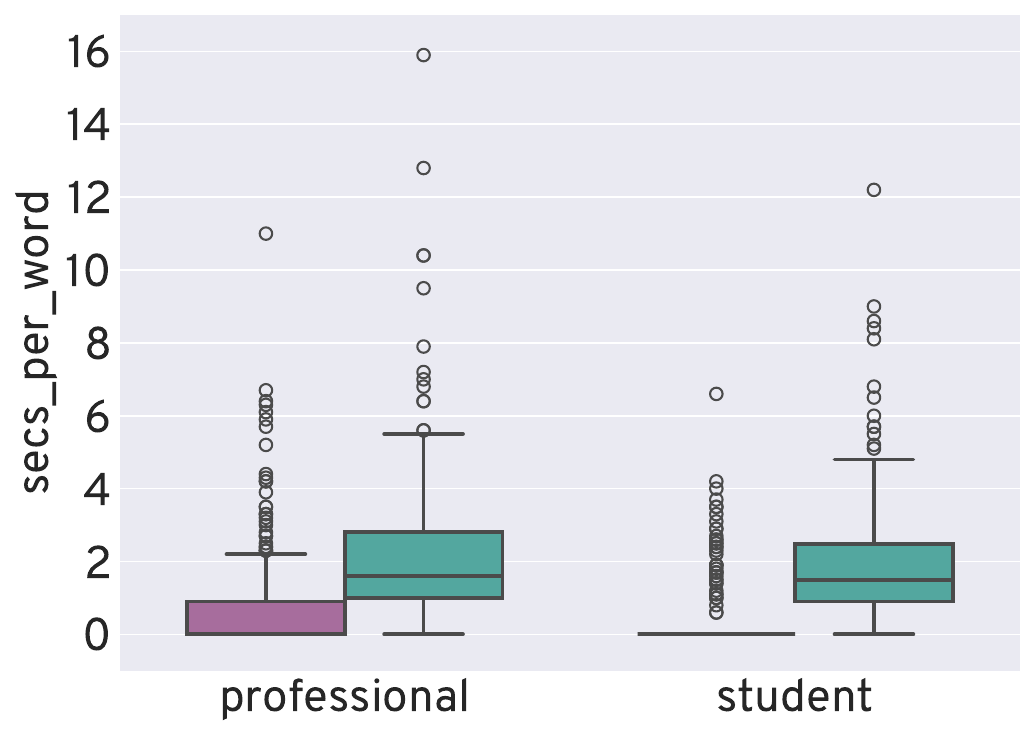}
        \caption{Multiuser}
        \label{fig:sec_multiuser_sent}
           \end{subfigure}
        
    \caption{Seconds per source word distribution across post-edited sentences.}
    \label{fig:sec_per_word_sent}
\end{figure*}

\paragraph{Data collection and effort measures}

At the end of the process, for each sample of 250 source sentences we collect 500 post-edits (250 F and 250 M). We then measure the corresponding ``feminine''/``masculine'' effort for the \textit{temporal} and \textit{technical} dimension \citep{krings2001repairing}. 
Respectively, \textit{i)} time to edit (TE) is recorded within Matecat for  each output sentence,\footnote{Sentences that do not require any post-editing count as 0.}  whereas \textit{ii)} the amount of edits is computed with HTER \citep{snover2006study}. 

We frame the difference between feminine and masculine efforts ($\Delta$) as our human-centered measure for gender-related quality of service
disparities. 
We also  
compute statistical significance tests
between F and M effort values. We use 
paired bootstrap resampling \citep{koehn-2004-statistical} for HTER, and Wilcoxon  \citep{Rey2011}\footnote{The Wilcoxon result is computed using \texttt{scipy 1.13.1}: \url{https://docs.scipy.org/doc/scipy/reference/generated/scipy.stats.wilcoxon.html}.} for both HTER and TE, with p-value $<0.05$. Tests were calculated for all results presented in the paper, and are all statistically significant.

\section{Results}
\label{sec:results}

\subsection{Multidataset Results}
\label{subsec:multidata}
In Table \ref{tab:absolute_results} 
(top),
we report the cumulative results for TE and the number of edits across genders for three en-it datasets. 
Consistently, though with variation across datasets,
\textbf{ our results confirm a 
significant effort difference across genders}.

The unambiguous \textsc{mtgen-un} exhibits the lowest gap, attesting that, when source text provides gender cues, GT can better handle 
feminine and masculine gender in the target language. Still, even in this context, F post-editing
amounts to a +36.3\% and +58.3\%  increase ($\Delta$\textit{rel}), respectively for TE and HTER. For the ambiguous datasets, the gap clearly widens. This is 
particularly 
notable for \textsc{mtgen-a}, which  -- compared to \textsc{must-she} -- presents a higher distribution of gendered words (see Table \ref{tab:data_stats}), which are also more prone to bias, i.e. professions. 
Compared to its M counterpart, F post-editing for this dataset requires around four times the effort both 
in time 
and
number of edits. 

Overall, effort distribution across post-edited sentences -- Figure \ref{fig:hter_multidataset_sent} for technical and \ref{fig:sec_multidataset_sent} for temporal effort -- attest that for the vast majority of M sentences, no post-editing at all was required.
This mirrors 
the known GT tendency to masculine default \citep{Piazzolla_Savoldi_Bentivogli_2023}.

Henceforth, we focus on the particularly biased \textsc{mtgen-a} sample\footnote{We choose \textsc{mtgen-a} also to include a non-romance language (de), since MusT-SHE is only available for en-it/fr/es.} for multilanguage and 
multiuser
comparisons.








\subsection{Multilanguage Results}
\label{subsec:multilang}
Moving onto multilanguage assessments with  \textsc{mtgen-a}, we attest that human-centered disparities are present also for en-de and en-es.
Although cumulative results in Table \ref{tab:absolute_results} 
(center) 
show some variation for TE -- especially for the masculine set -- sentence-level distributions for both types of effort are highly comparable.  
In figure \ref{fig:hter_multilang_sent}, median HTERs are the same for en-de/it in the feminine set (14.3), and slightly lower for en-es (12.5). For masculine PE, the median HTER values are systematically 0, although the number of not edited sentences is visibly higher for en-de.\footnote{
Based on a manual analysis,
this is due to a lower incidence of \textit{preferential edits} (i.e. not gender-related), suggesting that 
post-editors
perceived the en-de output as of high quality.}
Median temporal efforts based on the number of source words per second are also very close, i.e. always 0 for M; whereas in the feminine PE we find 1.6 (en-it) 1.2 (en-es) 1.1 (en-de)  -- see Figure \ref{fig:sec_multilang_sent}. 
Overall,
\textbf{\bs{differences in efforts based on gender generalize across the considered language pairs}}.



\subsection{Multiuser Results}
\label{subsec:multiuser}
As a last step, we confront the PE activity of professional translators (P) with less experienced 
high-school students (S). 
Cumulative results in Table \ref{tab:absolute_results} (bottom) show that \textbf{in the student condition gender gaps widen significantly}. More specifically, 
percentage differences
for \textsc{mtgen-a} en-it 
go from +177.6\% (TE) and  +201.8\% (HTER) -- assessed with professionals -- up to respectively +329.8\% and +636.3\% for students. Quite surprisingly, and also confirmed by the distributions in Figures \ref{fig:hter_multiuser_sent} and \ref{fig:sec_multiuser_sent}, students are overall quicker, and edit less across both F and M.

We explain these results 
by the lower familiarity with both the English language and the PE task itself.
In fact, based on 
observations
during the 
experiments, also confirmed by manual revision of the collected post-edits,
students did not engage with the improvement of the overall accuracy of the translation. Rather, they almost exclusively focused  
on 
adjusting
gendered words.\footnote{Post-editing examples available in Appendix \ref{app:post-editing}.} 
Thus, to a certain extent, students' results allow us 
to isolate even more neatly
the sole effect of gender bias in MT with our human-centered measurements, an issue that might be further amplified should lay users be involved in similar experiments.





\section{Discussion}
\label{sec:discussion}

We found strong evidence for the human-centered impact of bias in MT, with a quality of service disparity that can disproportionately affect women.
Such allocative harm is evident in the extra time and energy required for feminine gender translation. 
Note that our results are likely conservative, involving experienced users with high language proficiency. Indeed, in less controlled conditions, or among individuals with lower proficiency in either 
target  or
source language, such a negative impact would likely be even greater. Misgendered references may go unnoticed, propagating errors in texts and communications, or necessitating the use of external resources such as dictionaries to be fixed. Due to experimental constraints (\S\ref{subsec:settings}), such a scenario remains open to future analyses. 
To better frame the implications of our findings, we conclude with two critical reflections. First, individuals might rely on third-party language services to translate their text, thus raising the question: \textit{Can gender bias imply a differential in economic cost}? Second, while informative assessments that center users are crucial to guide the field forward, \textit{are current automatic evaluations able to capture such human-centered disparities}?




\paragraph{Someone has to pay for the cost of gender bias.} 
We explore the economic costs of F and M post-editing considering two stakeholders: \textit{i)} a \textit{final user}, who buys the PE 
text from  
\textit{ii)} a
third-party \textit{translator}. As a case study, we analyze the three en-it datasets edited by professionals (\S\ref{subsec:multidata}) --
using averaged HTER and source words shown in Table \ref{tab:costs}. 
Note that pricing in the language industry is complex \citep{lambert2022because}
and can be based on various parameters \citep{scansani2020lsps, ginovart2020report}.
Here, we consider two common payment scenarios -- i.e. \texttt{HTER-Rate} and \texttt{Word-Rate}. 
For both payments,
we use a baseline word-rate of €0.09 per source word, reflecting 
best  
market prices for en-it \citep{inboxtranslation2023freelance}. 

\texttt{HTER Rate}:
With this method, prices are adjusted based on the \textit{actual} technical effort required to 
post-edit,
with lower edit rates leading to lower costs, and vice versa.
Following existing price schemes \citep{localization2022fair},\footnote{See  Figure \ref{fig:pricing} in  Appendix \ref{app:payment}.} HTER  below 10 is paid at 35\% of the word rate (i.e. €0.0315 per word), whereas HTER between 10-20 is paid at 40\%  (i.e. €0.036 per word). Hence, and as shown in Table \ref{tab:costs} (HTER€), feminine PE would cost more. 
While translators are compensated for the additional effort, such a financial burden will inevitably fall on the final user purchasing the F translation.

\begin{table}[t]
\footnotesize
\centering
\begin{tabular}{lcc|ccc}
\toprule
     & HTER  & src-W      & \textbf{HTER€}  & \textbf{Word€}                \\
    \midrule
\textsc{Fem}  & 10.92 & 5629     & 202.63 & 177.30 \\
\textsc{Mas} & 4.60  & 5629      & 177.30 & 177.30 \\
\bottomrule
\end{tabular}
\caption{Economic costs of feminine and masculine en-it PE. We provide pricing based on technical effort (HTER€) and on source text lenght (Word€). }
\label{tab:costs}
\end{table}

\texttt{Word Rate}: This pricing is based on 
source text length, where the cost per word is 
decided  \textit{a priori}. 
For PE tasks, the word-rates 
vary depending on the content or the language \citep{sarti-etal-2022-divemt}.\footnote{e.g. creative texts or certain languages are notably poorly handled by MT, thus corresponding to higher word-rates.}
For en-it data from a general domain such as ours, a 35\% word rate could be paid. 
Given that -- to the best of our knowledge -- this type of pricing does not consider gendered content, the same word-rate would be indiscriminately applied to both feminine and masculine PE. 
Thus, as shown in Table \ref{tab:costs} (Word€), a final user buying their translation would pay the same price, regardless of gender.
However, this would place the financial cost on the translator, whose additional effort required for feminine PE would be underestimated and under-compensated.

\noindent\finding{} To sum up, this 
analysis shows that gender bias has an economic cost which can unfairly fall onto either one of the two PE stakeholders.
Besides financial implications, unfair compensation could also invite less edits 
than
necessary, thus
compromising the quality of feminine PE. 
Analysing such potential quality-oriented implications is a crucial aspect for future research.


\begin{figure*}[h!]
    \centering
    \begin{subfigure}[t]{0.28\textwidth}
            \centering
            \includegraphics[width=\linewidth, 
            ]{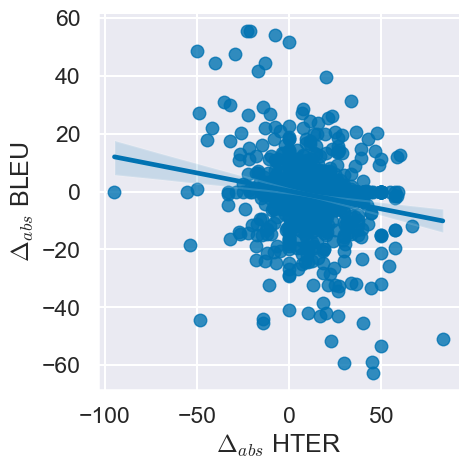}
            \caption{$\Delta_{abs}$ HTER and BLEU}
            \label{fig:cor_hter}
        \end{subfigure}
    \quad
    \begin{subfigure}[t]{0.28\textwidth}
            \centering
            \includegraphics[width=\linewidth, 
            ]{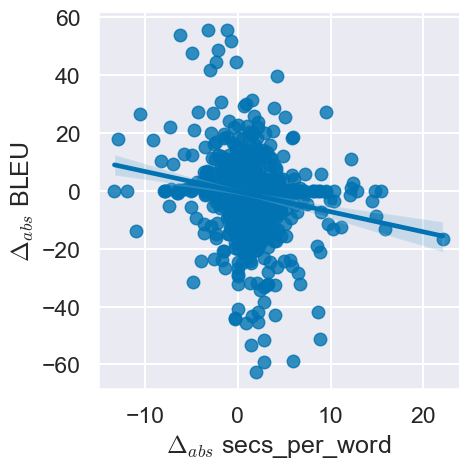}
            \caption{$\Delta_{abs}$ secs\_per\_word and  BLEU}
            \label{fig:cor_sec}

        \end{subfigure}
    \quad
    \begin{subfigure}[t]{0.28\textwidth}
            \centering
            \includegraphics[width=\linewidth, 
            ]{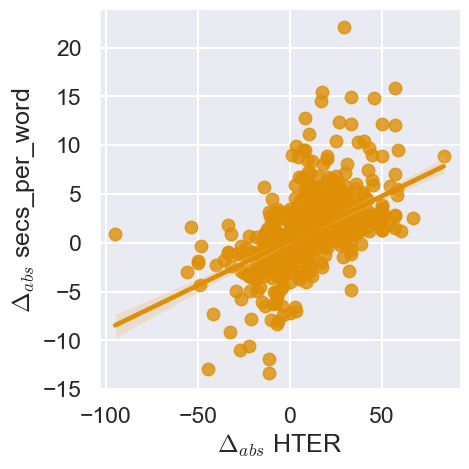}
            \caption{$\Delta_{abs}$ HTER and secs\_per\_word}
            \label{fig:cor_human}
        \end{subfigure}
    \caption{
    Scatter plots with overlaid regression lines of the differences between F and M scores for all \textit{datasets}, \textit{languages} and \textit{users}. Each point represents a sentence-level difference. The correlation between the different metrics is measured with the Pearson $r$ coefficient, and all results are statistically significant (p-value $<0.05$).\footnotemark}
    \label{fig:scatterplots}
\end{figure*}
\footnotetext{{Computed using \texttt{scipy 1.13.1}: \url{https://docs.scipy.org/doc/scipy/reference/generated/scipy.stats.pearsonr.html}}}

\paragraph{Automatic bias measurements do not reliably correlate with human-centered measures.}
Methods to quantify bias are key to 
much research that seeks to monitor
the 
creation of 
equitable technologies \citep{dev2022measures}.  
In this context, growing evidence underscored how \textit{intrinsic} metrics---focusing on models' representations---might not be a reliable bias indicator in downstream, real-world tasks, as assessed with \textit{extrinsic} ones---focusing on models' output \citep{jin-etal-2021-transferability, goldfarb-tarrant-etal-2021-intrinsic, cao-etal-2022-intrinsic, orgad-belinkov-2022-choose}.  
Arguably, however, even extrinsic measures are model-centric (\S\ref{sec:review}), and only assumed to reflect more
reliably 
the downstream harms to individuals. 
We verify this assumption by comparing  
our 
human-centered measures of 
differential effort 
with the 
automatic evaluations 
associated with MTGenEval and MuST-SHE (\S\ref{subsec:settings}). 
As in the original papers, we use the 
set of contrastive F/M target references\footnote{e.g. \textit{I am a \textbf{friend}} $\rightarrow$ M-es: \textit{soy \textbf{amigo}}, F-es: \textit{soy \textbf{amiga}}.}
to compute gender-related performance differences with BLEU\footnote{\texttt{nrefs:1|case:mixed|eff:no|tok:13a|smooth:exp}} \citep{papineni-etal-2002-bleu},
i.e. \textit{BLEU$_{F}$ -- BLEU$_{M}$}. 
%
%
%
%
Scatter plots of the automatic 
(i.e., BLEU score) 
and human-centric metrics (i.e., HTER and TE) differences, in absolute values, are reported in Figure \ref{fig:scatterplots}.
We provide aggregate results for all languages, datasets and users.\footnote{We also compute separate statistics for each sample, and with other metrics (\bs{COMET-22} and TER). The hereby discussed trends are confirmed. See Appendix \ref{subsec:correlations}. Details on the automatic metrics computation are provided in Appendix \ref{app:automatic-metrics}.}  

Looking at our results, 
we notice 
a Pearson-$r$ of 
$-0.19$
for $\Delta_{abs}$ HTER and $\Delta_{abs}$ BLEU (Figure \ref{fig:cor_hter}), and 
$-0.18$
for $\Delta_{abs}$ secs\_per\_word and $\Delta_{abs}$ BLEU (Figure \ref{fig:cor_sec}). The negative correlation is expected since, while for BLEU the higher the better, the opposite is true for both HTER and TE. Still,  the results clearly indicate that both temporal and technical efforts are in \textit{weak} correlation  \citep{Schober2018CorrelationCA} with automatic scores. 
On the one hand, it is known that 
time measures 
may not always have a linear relationship with textual differences measured by automatic metrics \citep{tatsumi2009correlation, macken}, e.g. even small edits can require a high cognitive load and more time. 
On the other hand, given that both BLEU and HTER capture surface modifications and quantity of edits, their weak correlations are particularly noteworthy.\footnote{\bs{As a matter of fact, additional results reported in Appendix \ref{subsubsec:correlations-ter-comer} show that COMET -- despite its attested higher degree of correlation with human assessments for overall MT  quality -- exhibit a \textit{very weak} correlation with human-centered measures of bias.}} 
A  \textit{moderate} correlation (Person-$r$ 
$0.54$) is found only between the human-centered measures HTER and TE. As observed in Figure \ref{fig:cor_human}, the higher the number of edits, the more time required.


\noindent\finding{}\bs{Overall, our results suggest that existing \textit{model-centric} measures of gender bias in MT might not reliably reflect the downstream harms to users. While the contrastive evaluation approaches explored here have been used to reveal gender gaps \citep{bentivogli-etal-2020-gender, currey-etal-2022-mt}, they do not correlate with or accurately reflect the magnitude of disparities found through more concrete, human-centered measures.\footnote{See also the contrastive, automatic bias scores reported in Table \ref{tab:contrastive_results} in Appendix \ref{app:bias-scores}.} To ensure that advancements in the field prioritize impacted individuals, future research should explore both the metrics and the data used to compute them \citep{orgad-belinkov-2022-choose}. This includes investigating how automatic metrics relate to human-centered measures and how they can be translated into more transparent, user-relevant evaluations \citep{liao2023rethinking}.}




\section{Conclusion}


From cars' safety measures more effective for men, 
\citep{ulfarsson2004differences}, 
to virtual reality headsets that are too big to wear \citep{robertson2016building}, evidence of social and technological advances being less functional for women, or even harmful, abounds \citep{depope2024invisible}. While it is increasingly acknowledged that also language technologies can 
contribute to broader patterns of gender bias, still little is known about their tangible impact on people.
Our study represents a novel effort to empirically examine the implications of gender bias in MT with a human-centered perspective. Previous research has often inferred the downstream impact of bias based on automatic, model-centric scores. In contrast, we provide concrete empirical evidence showing that gender bias in MT leads to tangible service disparities, which can disproportionately affect women. 
%
%
Also, we quantify these disparities using measures that are more meaningful to impacted individuals, such as workload and economic costs.Our study advocates for an understanding of bias and its impact that centers on the actual users of this technology  to guide the field. To this aim,
we make our collected data and metadata publicly available for future studies on the topic.

\section{Limitations}
\label{sec:limitations}


\textbf{Experimental construct.}
To foreground the impact of gender bias, our study employs datasets that include at least one gender translation phenomenon per sentence. 
While these data more closely simulate our scenarios of interest like  the translation of biographies or CVs -- where human gender references are common -- in other contexts
such phenomena may be more sparse. 
Despite potential variations in bias magnitude across different types of text, however, our findings would not change: gender bias would simply be more 
difficult
to detect.
Also, while women would likely be the main target of 
bias-related
issues, the found costs and disparities could actually fall on anyone attempting to use feminine expressions, e.g. current attempts to avoid ``masculine default'' expressions for generic referents, and rather rely on generically intended feminine forms \citep{merkel2017only}. 
Overall,
since we rely on two widely recognized MT gender bias benchmarks, the density of gender phenomena in our study is actually the same density that is automatically evaluated 
with current bias metrics.

\textbf{MT system.} We prioritize the type of languages, participants and datasets as variables of interest over including MT system comparisons.
This choice is also motivated by the fact that gender bias is a widespread issue in generic MT models \citep{savoldi-etal-2023-test}, and attested with limited variation in commercial MT applications \citep{rescigno2020case, troles-schmid-2021-extending}.
Despite being a commercial system that can limit reproducibility, we pick Google Translate as it represents one of the most used MT engines by the public. Also, we exclude experiments based on instruction-tuned models such as ChatGPT given that the language industry as well as end-users mostly rely on standard MT for core translation tasks \citep{Fishkin_ChatGPT_2023}.\footnote{This was also confirmed by our study participants.} Also, 
while ``gender-specific translation prompts'' could help in the future \citep{sánchez2024genderspecific}, they are currently less realistic as they require users to craft them and -- before that -- to be aware of the presence, and thus the need to control for gender bias in MT.

\textbf{Languages.} 
Our study focuses on the translation of English sentences into grammatical gender languages that distinguish between masculine and feminine forms to express the gender of human referents \citep{gygax}. As such, we should be cautious in generalizing our findings to languages that mark gender differently, or not at all. Also, we focus on three language pairs (en-it/es/de) that are well-supported by current MT. Hence, it remains open to future investigation 
if the human-centered impact of gender bias could vary for 
languages with overall lower MT quality. 

\textbf{ACL query.}  The review of prior work on gender (bias) in MT considers only literature from the ACL Anthology. While searching other sources could have enriched our analysis, the Anthology represents the main historical reference point in the field and serves as a good and sufficiently comprehensive litmus test for examining the main trends in NLP.

Finally, we discuss the limitations of our binary gender setup in the upcoming section. 



\section{Ethical Statement}
\label{sec:ethical}

Our study is limited to binary, \textit{feminine} and \textit{masculine}, linguistic expressions of gender. Indeed, this choice, as well as the use of gender as a variable, warrants some ethical reflections. 
First of all,  we stress that – by working on binary linguistic forms – we do not imply a binary vision on the extra-linguistic reality of gender and gender identities \citep{d2023data}. 
The motivation behind our binary design  was to ensure comparable conditions between gendered post-edits. While non-binary forms and neutral expressions are increasingly emerging in the target languages of our study \citep{spanish, mirabella2024role, daems-2023-gender, piergentili-etal-2024-enhancing}, 
the attitude towards these forms, as well as  their level of usage can  widely vary among speakers
\citep{spanish, piergentili-etal-2023-hi}.
Given that non-binary and neutral expressions are not standardized like masculine and feminine terms, incorporating them would necessitate controlling for participants' prior familiarity with these forms. 
This additional variable could introduce cognitive effort complicating the measurement of post-editing effort. By focusing solely on binary gender expressions, we aim to isolate the effort and costs that are exclusively due to the system's bias without confounding it with the potential cognitive load associated with producing non binary language \citep{lardelli-gromann-2023-gender, paolucci-etal-2023-gender}. While by all means of utmost importance for future research, we were not able for the time being to also account for this cognitive dimension, which would have required additional tools and costs.

\section*{Acknowledgements}

Beatrice Savoldi is supported by the PNRR project FAIR -  Future AI Research (PE00000013),  under the NRRP MUR program funded by the NextGenerationEU. 
The work presented in this paper is also funded by the Horizon Europe research and innovation programme,  under grant agreement No 101135798, project Meetween (My Personal AI Mediator for Virtual MEETtings BetWEEN People), and the ERC Consolidator Grant No 101086819. 
This research was made possible by the participation of several bodies and individuals that took part in our human-centered study. 
We 
thank
the Directorate-General for Translation (DGT) of the European Commission 
and the DGT translators that kindly agreed to participate in the activity for en-it. 
We also tank
the independent professional translators that worked with us across all language pairs, as well as the high-school students that participated in our laboratories, thus contributing to the multiuser experiments.
Finally, we thank Jasmijn Bastings for the insightful discussion on and contribution to the gender bias papers' review.

\bibliography{anthology,custom, anthology_p2}
\bibliographystyle{acl_natbib}

\appendix

\section{Details on ACL Anthology Search}
\label{app:acl_query}
Our ACL search is based on the combination of keywords displayed in Table \ref{tab:query_acl}. Note that we also include terms such as ``rewriters''', which several works apply to the output of MT models as a bias mitigation strategy to offer double feminine and masculine outputs. 
%
%
%
To avoid retrieving unrelated works that only marginally mentioned \textit{MT} or \textit{gender} in the main body, 
the searches parsed only the title and abstract of the queried papers.

\begin{table}[htp]
\footnotesize
\centering
\begin{tabular}{lll}
\toprule
& \textbf{Keywords} & \# Papers \\
\midrule
 main & \begin{tabular}[c]{@{}l@{}}translation, NMT, \\ MT, rewriter\end{tabular}\\
\cmidrule{2-3} 
+                 & gender &   138 \\
\cmidrule{2-3} 
+                 & bias & 113 \\
\cmidrule{2-3} 
+ +                & \begin{tabular}[c]{@{}l@{}}manual, survey,  human, \\ participant,  expert, \\ qualitative,  user,  \\ people,   annotat*, \\ linguist, professional\end{tabular} & 96 \\
\bottomrule
\end{tabular}
\caption{Number of search results for each specific keyword combinations on the ACL anthology. In total, we find 347 results comprising 251 unique articles, of which \bs{146} were discarded as out of scope.}
\label{tab:query_acl}
\end{table}

\paragraph{Manual selection}
We retrieved a total of 251 unique articles. 
Of those, we discarded all unrelated papers that refer to e.g. \textit{inductive bias}, \textit{bias lenght}, or "\textit{translation}", but not in relation to the MT task.
We thus arrive at a total of \bs{105} papers.
The whole selection was carried out manually, \bs{and we annotated both papers that that matched the query focusing on human assessment as well as those that did not, so not ensure not to overlook any paper involving humans. }
We defined the papers to be considered \textit{in-scope} as follows: 
\begin{itemize}
    \item \textbf{MT application}: we only keep those works that primarily focused on MT, whereas those that relied on MT as an intermediate tool (e.g. to automatically translate a set of data) are discarded.\footnote{Two papers  \citet{daems-2023-gender, paolucci-etal-2023-gender} that focused on gender (bias) translation, but did not focus on MT were discarded, too.} 
    \item \textbf{Modality}: while limited in number, we keep also MT beyond the text-to-text modality.
    \item \textbf{Gender (bias)}: we include in our selection all works that focus on gender translation in the context of human entities. This includes works that do not explicitly engage with the notion of social bias -- especially prior to 2018. Papers more broadly addressing gender fairness and inclusivity are also included. 
\end{itemize}

The full list of extracted papers that made our final selection is provided below.



\bs{The first \textit{in-scope}
papers date back to 2016, whereas the latest two are from 2024. As of April, in fact, only few papers had been included in the Anthology. These 2024 papers are thus not shown in the figure to avoid incomplete views on approaches for the present year.
}

\paragraph{MT gender bias papers, no human assessment}

\citet{van-der-wees-etal-2016-measuring};
\citet{rabinovich-etal-2017-personalized};
\citet{bawden-2017-machine-translation};
\citet{popel-2018-cuni};
\citet{michel-neubig-2018-extreme};
\citet{vanmassenhove-etal-2018-getting};
\citet{moryossef-etal-2019-filling};
\citet{escude-font-costa-jussa-2019-equalizing};
\citet{cho-etal-2019-measuring};
\citet{habash-etal-2019-automatic};
\citet{stafanovics-etal-2020-mitigating};
\citet{basta-etal-2020-towards};
\citet{costa-jussa-de-jorge-2020-fine};
\citet{saunders-etal-2020-neural};
\citet{gonen-webster-2020-automatically};
\citet{stojanovski-etal-2020-contracat};
\citet{rescigno-etal-2020-case};
\citet{bentivogli-etal-2020-gender};
\citet{saunders-byrne-2020-reducing};
\citet{hovy-etal-2020-sound};
\citet{gonzalez-etal-2020-type};
\citet{costa-jussa-etal-2020-gebiotoolkit};
\citet{troles-schmid-2021-extending};
\citet{savoldi-etal-2021-gender};
\citet{wisniewski-etal-2021-biais};
\citet{ciora-etal-2021-examining};
\citet{escolano-etal-2021-multi};
\citet{ramesh-etal-2021-evaluating};
\citet{levy-etal-2021-collecting-large};
\citet{gaido-etal-2021-split};
\citet{vanmassenhove-etal-2021-machine};
\citet{vincent-2021-towards};
\citet{renduchintala-etal-2021-gender};
\citet{castilho-etal-2021-dela};
\citet{wisniewski-etal-2021-screening};
\citet{vanmassenhove-monti-2021-gender};
\citet{wisniewski-etal-2022-biais};
\citet{costa-jussa-etal-2022-evaluating};
\citet{castilho-2022-much};
\citet{gete-etal-2022-tando};
\citet{solmundsdottir-etal-2022-mean};
\citet{savoldi-etal-2022-dynamics};
\citet{mechura-2022-taxonomy};
\citet{corral-saralegi-2022-gender};
\citet{mohammadshahi-etal-2022-compressed};
\citet{saunders-etal-2022-first};
\citet{karpinska-etal-2022-demetr};
\citet{zhu-etal-2022-flux};
\citet{sharma-etal-2022-sensitive};
\citet{wisniewski-etal-2022-analyzing};
\citet{vincent-etal-2022-controlling-extra};
\citet{wang-etal-2022-measuring};
\citet{renduchintala-williams-2022-investigating};
\citet{alrowili-shanker-2022-generative};
\citet{alhafni-etal-2022-arabic};
\citet{gete-etchegoyhen-2023-evaluation};
\citet{dinh-niehues-2023-perturbation};
\citet{singh-2023-gender};
\citet{iluz-etal-2023-exploring};
\citet{alhafni-etal-2023-user};
\citet{sandoval-etal-2023-rose};
\citet{wicks-post-2023-identifying};
\citet{piergentili-etal-2023-gender};
\citet{saunders-olsen-2023-gender};
\citet{kostikova-etal-2023-adaptive};
\citet{cabrera-niehues-2023-gender};
\citet{fucci-etal-2023-integrating};
\citet{lu-etal-2023-reducing};
\citet{castilho-etal-2023-online};
\citet{paulo-etal-2023-context};
\citet{le-etal-2023-challenges};
\citet{sarti-etal-2023-ramp};
\citet{vincent-etal-2023-mtcue};
\citet{costa-jussa-etal-2023-multilingual};
\citet{attanasio-etal-2023-tale};
\citet{lee-etal-2023-target};
\citet{wang-etal-2023-survey};
\citet{veloso-etal-2023-rewriting};
\citet{sarti-etal-2023-inseq}

\paragraph{MT gender bias papers, manual evaluation}

\citet{bawden-etal-2016-investigating};
\citet{stanovsky-etal-2019-evaluating};
\citet{gaido-etal-2020-breeding};
\citet{kocmi-etal-2020-gender};
\citet{caglayan-etal-2020-simultaneous};
\citet{choubey-etal-2021-gfst};
\citet{popovic-2021-agree};
\citet{jain-etal-2021-generating};
\citet{vamvas-sennrich-2021-contrastive};
\citet{vanmassenhove-etal-2021-neutral};
\citet{currey-etal-2022-mt};
\citet{wairagala-2022-gender};
\citet{savoldi-etal-2022-morphosyntactic};
\citet{alhafni-etal-2022-user};
\citet{savoldi-etal-2023-test};
\citet{triboulet-bouillon-2023-evaluating};
\citet{costa-jussa-etal-2023-toxicity};
\citet{soler-uguet-etal-2023-enhancing};
 \citet{savoldi-etal-2024-prompt};
\citet{liu-niehues-2024-transferable};

\paragraph{MT gender bias papers, survey}
\citet{daems-hackenbuchner-2022-debiasbyus}
\citep{lardelli-gromann-2023-gender};
\citet{piergentili-etal-2023-hi};
\citet{lauscher-etal-2023-em};
\citet{amrhein-etal-2023-exploiting}

\paragraph{MT gender bias papers, participatory}
\citet{gromann-etal-2023-participatory}


\section{Experimental details}
\label{app:experimental}

\subsection{Data Details}
Here we provide additional information concerning the selection of the data used in our experiments (\S\ref{app:data_info}). Also, some minor corrections were made on the \textsc{mtgen-a} reference translation (\S\ref{app:ref}).

\subsubsection{Data selection}
\label{app:data_info}

\paragraph{MTGenEval-A selection} The 250 sentences used in our en-it experiments represent a randomly selected sample of the ``ambiguous'' section of the original MTGenEval dataset \citep{currey-etal-2022-mt}.
For the multilanguage experiments, we also maximize the overlap between en-it/es/de subsets. 
Overall, we retrieve 76 sentences which are common across all languages, whereas the remaining are randomly extracted within each monolingual portion of the original dataset. 

\paragraph{MTGenEval-UN selection}
The \textsc{mtgen-un} sample used in our experiments  was randomly extracted from the  ``unambiguous'' section of the original MTGenEval corpus. Note that, by being a subset with unambiguous gender in the English source, for this sample we extract 250 \textit{pairs} of sentences, for a total of 500. To exemplify, each pair corresponds to \textit{i)} a \textit{feminine} <source-target> segment (e.g. en: ``Sarandon has \textbf{appeared} in two episodes of The Simpsons, once as \underline{herself} and...'', it: ``Sarandon è \textbf{apparsa} in due episodi dei Simpson, una volta interpretando se \textbf{stessa}...''), and \textit{ii)} a \textit{masculine} <source-target> segment (e.g. en: ``Sarandon has \textbf{appeared} in two episodes of The Simpsons, once as \underline{himself} and...'', it: ``Sarandon è \textbf{apparso} in due episodi dei Simpson, una volta interpretando se \textbf{stesso}...''). We automatically translate with GT the total 500 English sentences and create the corresponding feminine and masculine samples of 250 sentences each to be post-edited. 

\paragraph{MuST-SHE selection}
For \textsc{must-she}, which by design contains an higher variety of gender phenomena for several parts of speech 
we 
relied on preliminary filters to ensure a less noisy experimental environment. Namely, we excluded sentences that in the original dataset are annotated as ``FREE-REF'', and for which the human reference translation is known to be quite creative and less literal. Also, prior work based on this dataset has shown that -- due to its higher variability -- a good amount of gendered words available in the reference translation might not be actually generated in the MT output for a range of reasons, i.e.  errors, synonyms etc \citep{savoldi-etal-2022-morphosyntactic}. 
Thus, first we translated the whole corpus with Google Translate. Then, we only retained those sentences where the MT output  contained at least one gendered word annotated in the corresponding reference translations. To do so,  we relied on the \textit{coverage} evaluation script\footnote{\url{https://github.com/hlt-mt/FBK-fairseq/blob/master/examples/speech_to_text/scripts/gender/mustshe_gender_accuracy.py}} made available with the corpus. Overall, these filters ensured \textit{i)} the presence of gender phenomena to revise during the PE task, \textit{ii)} less creative reference translation that eased more reliable assessments with automatic metrics. The final 250 sentences were randomly extracted from this pre-filtered MuST-SHE subset.


\subsubsection{MTGenEval-A reference translations}
\label{app:ref}

For \textsc{mtgen-a}, we find that for some English sentences not all ambiguous human entities are 
translated
with masculine or feminine gender in the 
corresponding reference of the M/F contrastive pair.
We thus manually revised all reference translations for for the 3 en-it/es/de datasets. This is necessary to align the results of our PE activity -- where \textit{all}  entities whose gender is ambiguous in English 
are post-edited either as masculine or feminine -- with the automatic bias evaluation method presented in Section \ref{sec:discussion}, which is based on the reference translations.
To exemplify, see the following en-es segment:  

\smallskip

\noindent \textit{src-en}:  \textbf{The doctor} and \textbf{some} of \textbf{the} patients had signed off to purchase it

\noindent \textit{tgt-es-F}:
\textbf{La doctora} y \textbf{algunos} de \textbf{los} pacientes se habían apuntado para comprarlo.	

\noindent \textit{tgt-es-M}: \textbf{El doctor} y \textbf{algunos} de \textbf{los} pacientes se habían apuntado para comprarlo.

\smallskip

While ``doctor'' is respectively 
translated as masculine or feminine in the 
corresponding references, the equally ambiguos ``some of the patients'' is not, and rather remains masculine in both references. To fix these instances, 
for each of the 250 source sentences included in the en-it, en-es and en-de datasets, 
we manually revised both reference translations.

This was carried out by a linguist with expertise in all languages pairs. Overall, 40 segments were modified for en-it, 15  for en-es, and 28 for en-de.

\subsection{Matecat tool and settings}
\label{app:matecat}

To work in Matecat,\footnote{\url{https://www.matecat.com/}} 
we created two separate projects for each dataset: one for the feminine setting and one for the masculine setting.  
For each project, we followed the same procedure. 
We uploaded the input English text 
and created a 
corresponding dedicated 
Translation Memory (TMX). 
The TMX contains the translations produced by GT, which are shown to the translators as suggestions to post-edit. 
Crucially, we ensured our settings as follows: \textit{i)} each 
translator
had access to the dedicated TMX in a ``lookup-only'' mode, meaning that they could not update it with their post-edits
-- which would have otherwise become visible to the other translators 
and make the experiment ineffective;
also, \textit{ii)} the general Matecat TMX was disabled, so as to avoid that translators had access to additional suggestions other than the GT outputs.;
then, \textit{iii)} to ensure that the Matecat tool would maintain the original sentence division of the dataset,
we activated the \textit{paragraph} setting, which does not re-segment the input text.
Finally, each M/F project was split into sub-projects of around 15 sentences each to be assigned to participants (14 splits for \textsc{mtgen-a}, 16 for \textsc{mtgen-un}, and 16 for \textsc{must-she}). Each participant received two links to work on both an M and an F sub-project, for a total of around 30 sentences to post-edit.

\subsection{Automatic Metrics}
\label{app:automatic-metrics}
The automatic metrics used to evaluate translation quality are BLEU, \citep{papineni-etal-2002-bleu}, based on n-gram matching, TER \citep{olive2005global}, based on edit rates, and the neural-based COMET \citep{rei-etal-2020-comet}. BLEU and TER are computed with the well-established tool for evaluating machine translation outputs, sacrebleu v2.4.0 \citep{post-2018-call}.\footnote{\texttt{nrefs:1|case:mixed|eff:no|tok:13a|smooth:exp}}
COMET is computed using the official GitHub repository\footnote{\url{https://github.com/Unbabel/COMET}} with the \texttt{Unbabel/wmt22-comet-da}\footnote{\url{https://huggingface.co/Unbabel/wmt22-comet-da}} model.


\begin{figure}[t]
    \centering
       
    \begin{subfigure}[t]{0.475\textwidth}
        \includegraphics[width=\textwidth]{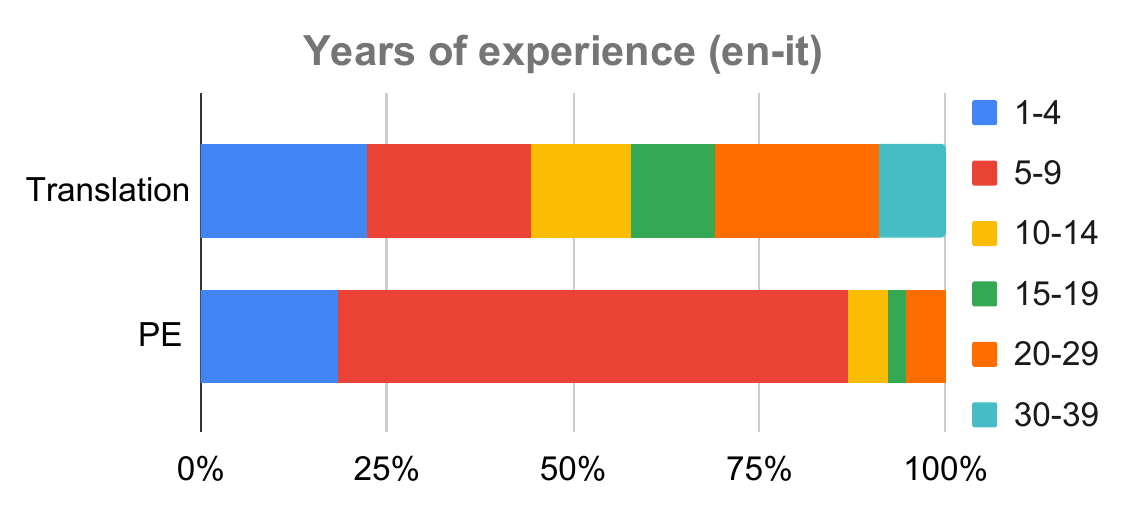}
    \end{subfigure}\\  
    \begin{subfigure}[t]{0.475\textwidth}
        \includegraphics[width=\textwidth]{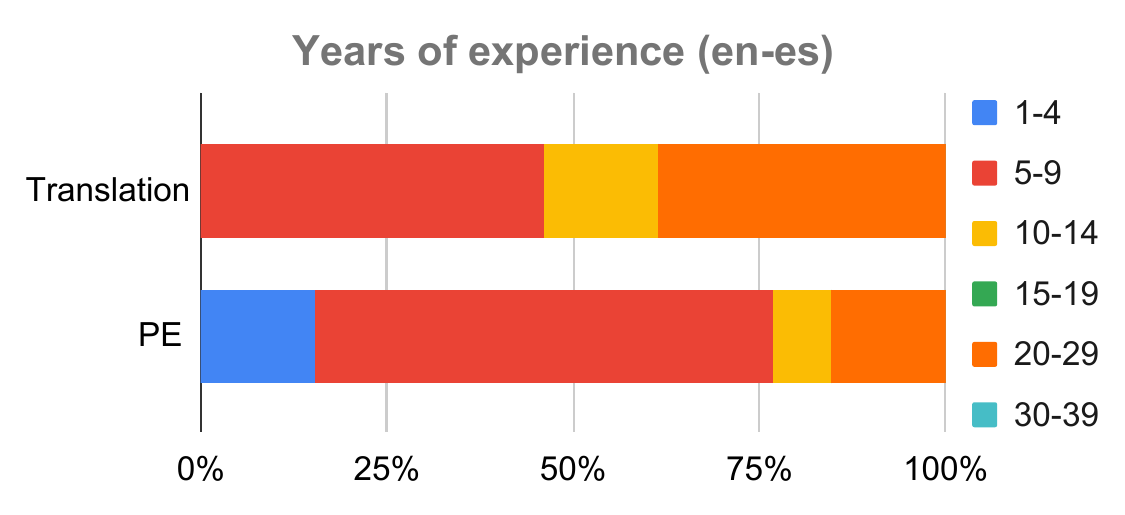}
    \end{subfigure}\\
    \begin{subfigure}[t]{0.475\textwidth}
        \includegraphics[width=\textwidth]{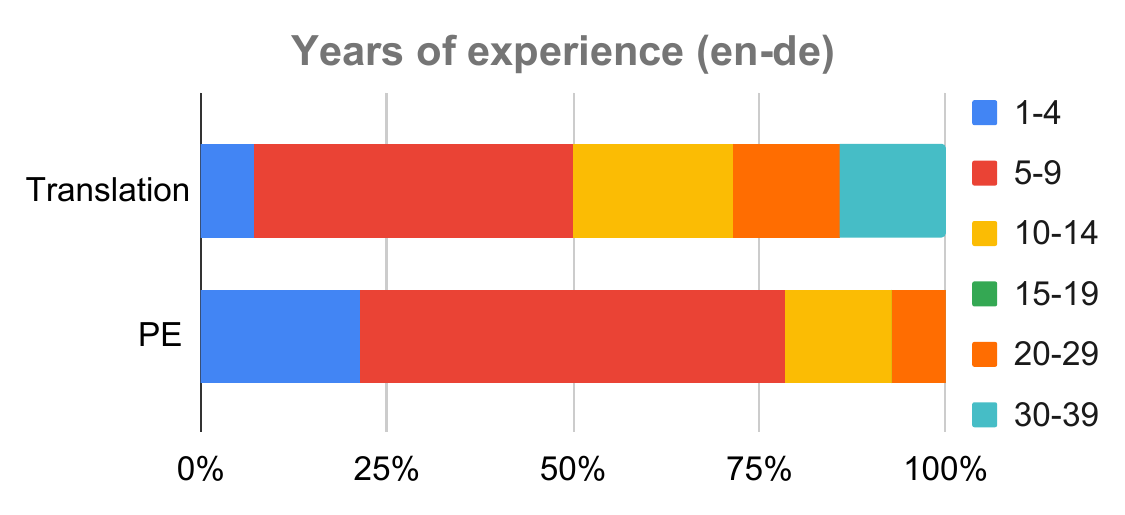}
           \end{subfigure}
        
    \caption{Professional translators' years of experience as translators, and as MT post-editors. Results are shown for each language pair.}
    \label{fig:participants}
\end{figure}

\section{Study participants}
\label{app:participants}

We relied on two types of participants in our experiments: \textit{professional translators} and \textit{high school students}.
As for 
translators, the experiment 
include professionals who 
participated
on a voluntary basis
as well as paid professionals.
To ensure comparability, we replicated the same settings and used the same guidelines across all conditions. 
For students, we 
added a warm-up phase to introduce them to MT, the PE task, and the Matecat tool. 
%

All the experiments were agreed upon with all participants. 
The\textbf{ privacy protection} of the involved participants is guaranteed by the complete anonymity of the whole collected data, which make it impossible to identify the involved subjects. 


\subsection{Recruitment and Task organization}
\label{app:recruitment}

\paragraph{Professional translators (volunteers)} For en-it, a first round of experiments was
carried out with professional 
translators from the European Commission, Directorate-General for Translation, Italian-language Department.
These participated on a voluntary basis as part of an educational lab 
held by the authors of this paper.
As such, no compensation was involved.

\begin{table*}[t]
\footnotesize
\centering
\begin{tabular}{l|cccc||cccc}
\toprule
 & \multicolumn{4}{c||}{\textbf{TE}}                           & \multicolumn{4}{c}{\textbf{HTER}}                \\
 & \textsc{Fem} & \textsc{Masc} & $\Delta$\textit{abs} & $\Delta$\textit{rel}  & \textsc{Fem} & \textsc{Masc} & $\Delta$\textit{abs} & $\Delta$\textit{rel} \\
 \midrule
\textsc{Voluntary Professionals}            
& 1:13 & 0:27 
& \cellcolor{mygray}0:46 &  \cellcolor{mydarkgray}170.40
& 14.31  & 2.39 & \cellcolor{mygray}12.78   & \cellcolor{mydarkgray}259.23 \\

\midrule
\textsc{Paid Professionals}   
& 1:07 & 0:26 
& \cellcolor{mygray}0:40  & \cellcolor{mydarkgray}150.95
& 17.71 & 4.93  & \cellcolor{mygray}11.92  & \cellcolor{mydarkgray}498.74  \\



\bottomrule
\end{tabular}
\caption{Comparative post-editing results for 125 sentences en-it on \textsc{mtgen-a}, carried out by the group of voluntary professional translators and the second setting of paid professional translators. We provide time to edit (TE, i.e. hour:minutes) and HTER.}
\label{tab:agr-en-it}
\end{table*}

To carry out experiments on \textsc{mtgen-a}, \textsc{mtgen-un}, and \textsc{must-she}, we 
needed
data from 14 + 16 + 16 participants, respectively, for a total of 46 participants.  
However, eventually 22 blocks of sentences (corresponding to the activity of 11 participants) were not carried out or completed. This was due to several reasons: some expected participants were absent, others experienced internet connection problems that hindered them to properly carry out the PE activity, and one participant decided not to take part in the experiment. 
%
%
Thus, in order to complete our data collection, we resorted to paid professional translators.

\paragraph{Professional translators (paid)} The remaining en-it data and all en-es and en-de data were post-edited by paid professionals,
who were recruited via a translation agency.
The only eligibility criterion we required was that the en-* pair assigned to them represented one of their main language direction in their professional work, and that they were native speakers of the target language (i.e. the same working condition of  volunteers). The experiments where carried out via online meetings, in groups of around 8 translators.
To avoid introducing any confounding effect that could influence their PE work, all post-editors were requested to remain in the meeting for its entire duration of 50 minutes, and compensation was time-based.
The total cost (translation agency recruitment and translator's work) amounted to \texteuro50 per post-editor, taxes excluded.

The similarities of the work carried out by the two types of professional translators, verified as discussed in Appendix \ref{app:agreement}, allowed us to merge all en-it data coming from professionals and carry out aggregated dataset-level analyses.

\paragraph{Students (volunteers)}
The activity of the students was carried out during a laboratory as a part of their school activity. These students were from a school offering a foreign language specialization, thus ensuring that they had a good (B2 level) proficiency in English. 
They were all part of the same class, attending the penultimate year of high school. All the activities were allowed under the consensus of their school supervisor and under the supervision of their regular teachers. 
For this task setting, we also included a warm-up phase to introduce the students to MT, the PE task, and the Matecat tool before starting the experiments.

\subsubsection{Participant Statistics}
For each pair of languages, in Figure \ref{fig:participants} we provide the years of experience of the involved professionals, both as translators (i.e. translating from scratch) as well as MT post-editors. In line with overall statistics in field,\footnote{\url{https://www.linkedin.com/pulse/lets-talk-gender-equality-translation-industry-josephine-matser/}} women make up the majority of involved translators (77\%). 
We did not enforce balanced gender distributions in the recruitment process and did not deem the gender of the translators as a significant variable. Indeed, feminine and masculine lexical terms are equally standard, grammatical forms used to refer to human referents, which are part of the current language. \bs{This is also confirmed by prior work \citep{popovic-lapshinova-koltunski-2024-gender}, which did not find translator's gender to be an indicative factor in gender translation.}
Participants were only instructed to use them in translation according to the provided gender information for each sentence.  

No personal information was collected for students. 

\begin{figure*}[t]
    \centering
    \includegraphics[width=1\linewidth]{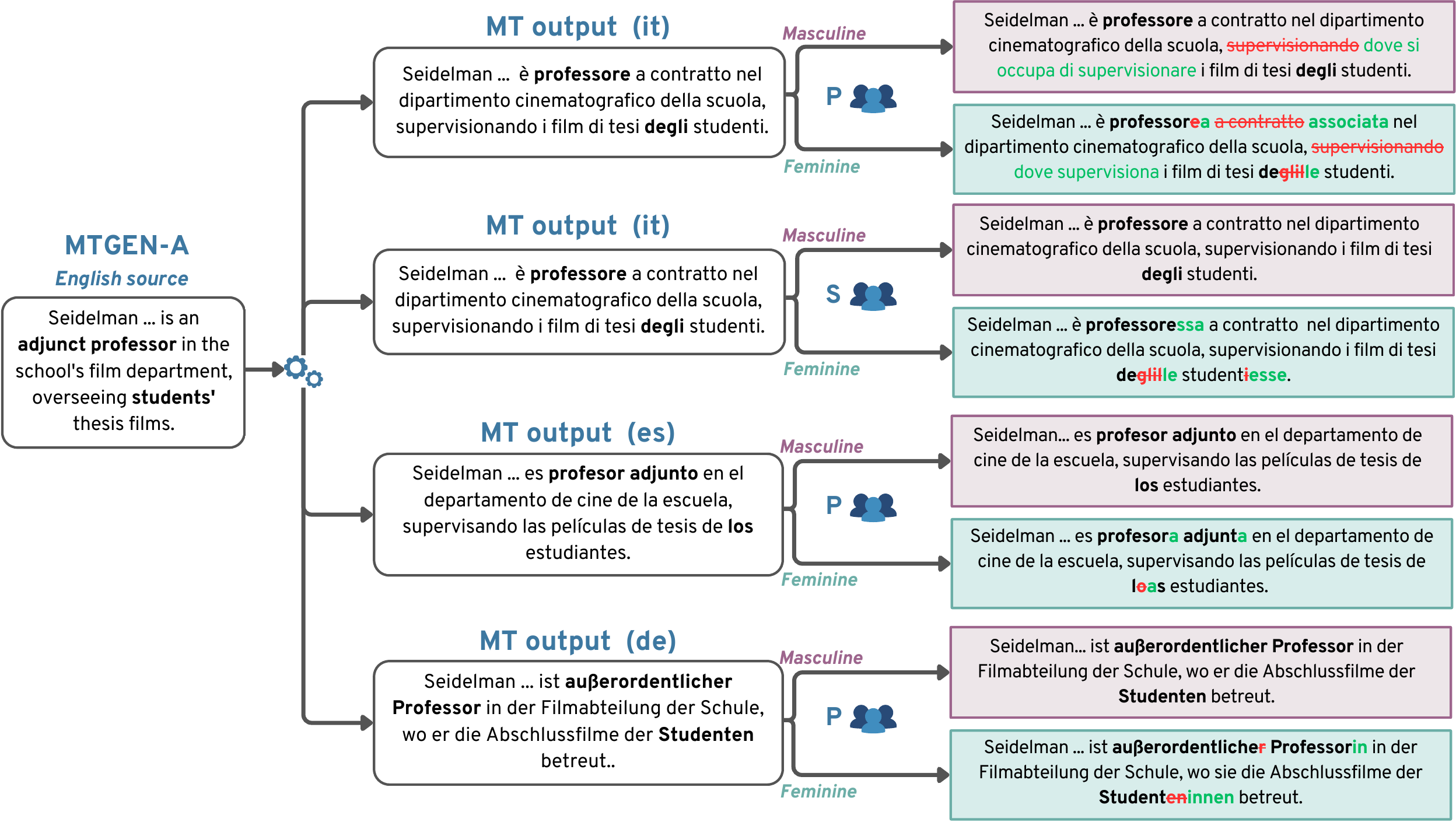}
    \caption{Post-editing example for a \textsc{mtgen-a} source English sentence, which is common across all language pairs. Given the source English sentence, we show the GT automatic translation, and its associated feminine and masculine post-edits. For en-it, we show post-editing by both professionals (P) and students (S). In bold, we show \textbf{gender-related words} in the source, output, and post-edited sentences. For the post-edits, we show \red{deletions} and \green{insertions}.}
    \label{fig:mtgena-PE}
\end{figure*}

\subsection{PE effort across voluntary and paid professionals}
\label{app:agreement}

Given that the PE activity for en-it is carried out by both paid and non-paid professionals (see Appendix \ref{app:recruitment}), we want to ensure that the two conditions are comparable. For this reason, we collected a \textit{control subset of sentences} -- edited by both paid professionals and voluntary professionals --  to compare the PE results across these two potentially different types of subjects. 
To do so, we have paid translators redo 125 sentences for \textsc{mtgen-a}, which is the dataset upon which most of our experiments are based. Hence, we collect an additional set of 300 post-edited sentences (i.e. the same 125 source sentences correspond to 125 F post-edits and 125 M post-edits).  

Results are reported in Table \ref{tab:agr-en-it}. 
As we can see, the type of professional does not appear as a significant confounding variable. In absolute numbers, the two sets are highly comparable, with only a 6-minute difference in TE, and less than 1 HTER score ($\Delta$\textit{abs}).

Given the results of this analysis, we could safely merge the data coming from both types of translators to compose the final en-it datasets.
For \textsc{mtgen-a}, the 125 common sentences that we decided to keep for the main 
experiments are those post-edited by the professional translators, so as to allow for higher comparability with the fully "paid" en-es/de data samples.  



\section{Post-editing}
\label{app:post-editing}

In Figure \ref{fig:mtgena-PE}, we show an example of the PE activity carried our for the \textsc{mtgen-a} dataset. We provide an  English sentence which is common to all language pairs, associated  with its corresponding GT output, and both masculine and feminine post-edits showing the PE activity. As we can see from the figure, all GT outputs consistently translate human referents with masculine gender forms, which are then adjusted for the feminine PE. 

\paragraph{Student and Professional PE}
Still in Figure \ref{fig:mtgena-PE}, we show a typical behavioural difference that we attest between types of users for en-it. Namely, between professional translators (P) and less experienced students (S). 
As discussed in \S\ref{subsec:multiuser}, we find that students post-edited less (i.e. lower number of edits and in less time) compared to professionals. 
As a matter of fact, students did not engage with the improvement of the overall quality of the sentence, most likely due to their lower English proficiency, and rather mainly looked at the Italian target to fix gendered translation. In fact, in the provided example (the en-it blocks at the top), the GT output provided a poor translation for "\textit{overseeing}" -- rendered as "supervisionando", which is suboptimal in terms of fluency, overall also impacting the adequacy and readability of the sentence. Indeed, for both feminine and masculine PE, professionals carried out a light post-editing that also ensured an alternative translation for that portion of the sentence, whereas it was overlooked by students. Overall, since the adjustments made by students were basically only gender-related, the attested gender disparities measured with HTER and TE become even more visible.

\section{HTER Payment Rates}
\label{app:payment}

\begin{figure}[t]
    \centering
    \includegraphics[width=1\linewidth]{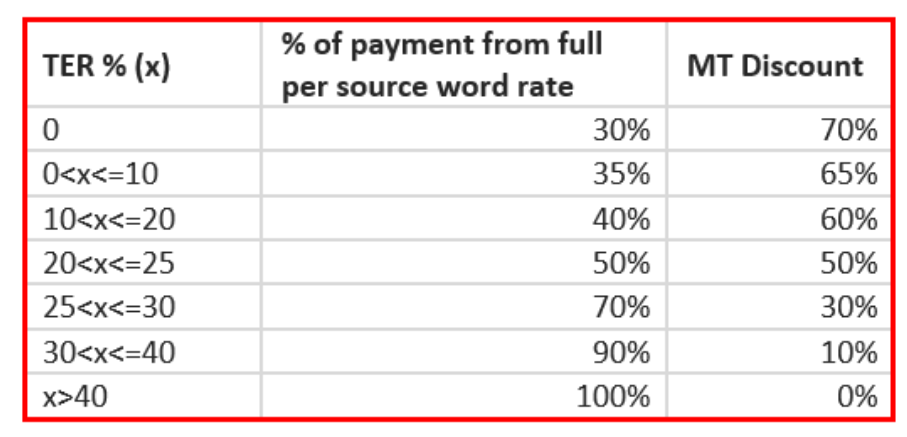}
    \caption{HTER Pricing matrix}
    \label{fig:pricing}
\end{figure}

To calculate HTER-based payments, we rely on the discount rates reported in Figure \ref{fig:pricing}. 
The matrix is publicly available and based on \citet{localization2022fair}. 
Note that discount rates can vary across companies. We compare the matrix with the HTER discounts used by other major language service providers.
Such rates however cannot be divulged as they are internal to the company and reserved. Overall, we find that the used scheme is highly aligned with those from other private companies and -- if anything -- it is more conservative, with a limited number of HTER ranges.

\section{Additional Results}
\label{app:additional_results}

\subsection{Overall Translation Quality}
\label{app:overall-translation-quality}

In Table \ref{tab:overall_quality} we report overall translation quality results obtained by Google Translate for all datasets and languages. We used the original target reference translation to compute the results. 

Details on automatic metrics computation are available in Appendix \ref{app:automatic-metrics}. 

\begin{table}[htpb]
\footnotesize
\setlength{\tabcolsep}{2.9pt}
\centering
\begin{tabular}{llccc}
\toprule
 &
   &
  \textbf{BLEU} ($ \uparrow $) &
  \textbf{TER} ($ \downarrow $) &
  \textbf{COMET} ($ \uparrow $) \\
\midrule
\textit{en-it} & \textsc{must-she}      & 40.64 & 47.54 & 84.56 \\
\textit{en-it} & \textsc{mtgen-un} & 43.92 &  42.92 & 82.31 \\
\cmidrule{2-5}
\textit{en-it} & \textsc{mtgen-a}  & 35.77 &  50.44 & 84.75 \\
\textit{en-es} & \textsc{mtgen-a}  & 49.72 &  34.2  & 85.29 \\
\textit{en-de} & \textsc{mtgen-a} & 36.04 &  49.35 & 84.28 \\
\bottomrule
\end{tabular}
\caption{Overall quality translation results per each dataset and language.}
\label{tab:overall_quality}
\end{table}









\subsection{Automatic  gender bias results}
\label{app:bias-scores}

We report contrastive, reference-based gender bias results computed with different metrics in Table \ref{tab:contrastive_results}. For  details on the metrics computation, please refer to Appendix \ref{app:automatic-metrics}. 

As expected, and in line with our post-editing results discussed in \S\ref{sec:results}, the unambiguous dataset \textsc{mtgen-un} obtains the smallest difference in scores. Overall, by looking at the differences in score computed against the feminine and masculine references ($\Delta$) also automatic evaluation methods confirm that GT exhibits gender bias, leading to a higher generation of masculine forms.  However, we immediately see that the \textit{magnitude} of such differences is notably small compared to our human-centered results reported in the main experiments of the paper (see \S\ref{sec:results}). This is particularly true for COMET, which is less sensitive to surface differences, such morphological gender-related differences. Overall, however, none of these metrics appear particularly sensitive at capturing gender differences, which are at best framed as +26.79 percentage difference as measured with TER (see \textsc{mtgen-a} for en-es). To further investigate this point, in the upcoming Appendix \ref{subsec:correlations} we verify the correlation between automatic scores and our human-centered measures.

\begin{table}[t]
\centering
\setlength{\tabcolsep}{1.5pt}
\footnotesize
\begin{tabular}{llcc|cc|cc}
\toprule
\multicolumn{1}{l}{} &
\multicolumn{1}{l}{} &
\multicolumn{2}{c}{\textbf{BLEU} ($ \uparrow $)} &
\multicolumn{2}{c}{\textbf{TER} ($ \downarrow $)} &
\multicolumn{2}{c}{\textbf{COMET} ($ \uparrow $)} \\
\toprule
\multicolumn{1}{l}{} & \multicolumn{1}{l}{} & \textsc{Fem} & \textsc{Mas} & \textsc{Fem} & \textsc{Mas} & \textsc{Fem} & \textsc{Mas} \\
\cmidrule{3-8}
\textit{en-it} & \textsc{must-she} & \cellcolor{fem}37.15 & \cellcolor{masc}43.51 & \cellcolor{fem}50.38 & \cellcolor{masc}45.18 & \cellcolor{fem}83.59 & \cellcolor{masc}85.43 \\
\textit{en-it} & \textsc{mtgen-un} & \cellcolor{fem}42.9 & \cellcolor{masc}44.94 & \cellcolor{fem}43.94 & \cellcolor{masc}41.91 & \cellcolor{fem}84.25 & \cellcolor{masc}84.86 \\
\textit{en-it } & \textsc{mtgen-a} & \cellcolor{fem}30.63 & \cellcolor{masc}39.8 & \cellcolor{fem}55.05 & \cellcolor{masc}46.79 & \cellcolor{fem}83.02 & \cellcolor{masc}86.52 \\
\textit{en-es} & \textsc{mtgen-a} & \cellcolor{fem}43.53 & \cellcolor{masc}54.56 & \cellcolor{fem}38.95 & \cellcolor{masc}30.72 & \cellcolor{fem}83.75 & \cellcolor{masc}86.64 \\
\textit{en-de } & \textsc{mtgen-a} & \cellcolor{fem}30.29 & \cellcolor{masc}40.52 & \cellcolor{fem}53.99 & \cellcolor{masc}45.87 & \cellcolor{fem}82.82 & \cellcolor{masc}85.77 \\
\toprule
\multicolumn{1}{l}{} &
\multicolumn{1}{l}{} &
$\Delta$\textit{abs} & $\Delta$\textit{rel} &
$\Delta$\textit{abs} & $\Delta$\textit{rel} &
$\Delta$\textit{abs} & $\Delta$\textit{rel} \\
\cmidrule{3-8}
\textit{en-it} & \textsc{must-she} & -6.36 & -14.62 & 5.20 & 11.51 & -1.84 & -2.15 \\
\textit{en-it} & \textsc{mtgen-un} & -2.04 & -4.54 & 2.03 & 4.84 & -0.60 & -0.71 \\
\textit{en-it } & \textsc{mtgen-a} & -9.17 & -23.04 & 8.26 & 17.65 & -3.50 & -4.05 \\
\textit{en-es } & \textsc{mtgen-a} & -11.03 & -20.22 & 8.23 & 26.79 & -2.89 & -3.34 \\
\textit{en-de} & \textsc{mtgen-a} & -10.23 & -25.25 & 8.12 & 17.70 & -2.95 & -3.44 \\
\bottomrule
\end{tabular}
\setcounter{table}{8}
\caption{Contrastive reference-based evaluation results for each language and dataset (Top), as computed with different metrics. Below, we show absolute difference ($\Delta$\textit{abs}) and percentage difference ($\Delta$\textit{rel}) values between feminine and masculine scores.}
\label{tab:contrastive_results}
\end{table}

\subsection{Correlation with automatic metrics}
\label{subsec:correlations}

\subsubsection{Aggregated results with COMET and TER scores}
\label{subsubsec:correlations-ter-comer}
As already discussed in Section \ref{sec:discussion}, performance differences in automatic metrics show a weak correlation with differences in human-centric metrics. This trend is reconfirmed by both COMET and TER scores, as shown in Figure \ref{fig:scatterplots_comet_ter}. Here, we still present aggregate results computed for all datasets, languages, and types of users. 

For the differences in COMET, we observe a relatively sparse distribution in Figure \ref{fig:scatterplots_comet_ter}.a, with a Pearson-$r$ coefficient of $-0.12$, meaning a \textit{very weak} negative correlation, against HTER. Similarly, the Pearson-$r$ coefficient against temporal effort (seconds per word) is $-0.17$, which is slightly higher but still represents a \textit{very weak} correlation. 
Even in the case of COMET, the correlation is negative because lower scores are better, while the opposite is true for HTER and sec\_per\_word. Moreover, when compared to Figure \ref{fig:scatterplots}, we observe a very similar behavior of BLEU (\S\ref{sec:discussion}, Figure \ref{fig:scatterplots}) with the one shown by COMET in Figure \ref{fig:scatterplots_comet_ter}.a and \ref{fig:scatterplots_comet_ter}.b, resembling similar distributions. Looking at TER differences, the samples of the distributions are slightly more squeezed towards the regression line. This means that the correlation is slightly higher but, however, still reaming \textit{very weak}, both considering HTER ($r=0.14$), and secs\_per\_word ($r=0.18$). In this case, the correlations are positive since the higher TER scores the better, similar to human-centric metrics.

\begin{figure}
    \centering
    \begin{subfigure}[t]{0.3\textwidth}
            \centering
            \includegraphics[width=\linewidth, 
            ]{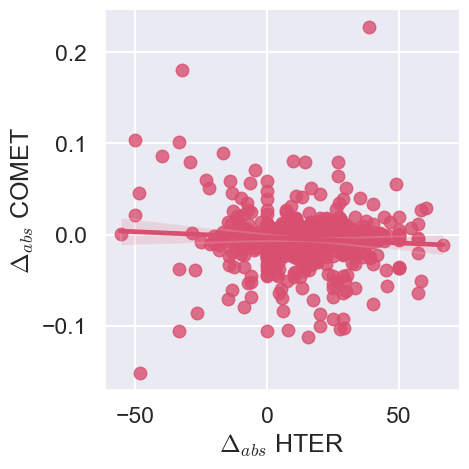}
            \caption{$\Delta_{abs}$ HTER and COMET}
        \end{subfigure}
    \qquad\qquad\qquad
    \begin{subfigure}[t]{0.3\textwidth}
            \centering
            \includegraphics[width=\linewidth, 
            ]{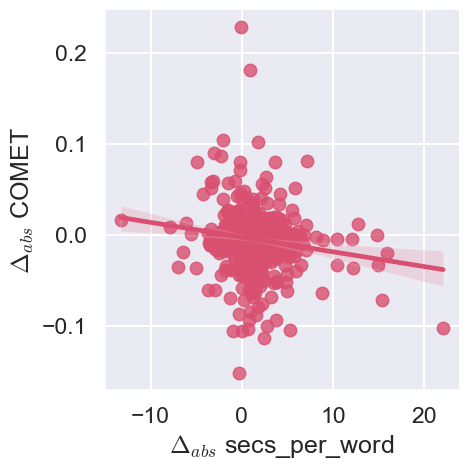}
            \caption{$\Delta_{abs}$ secs\_per\_word and COMET}
        \end{subfigure}\\
    \begin{subfigure}[t]{0.325\textwidth}
            \centering
            \includegraphics[width=\linewidth, 
            ]{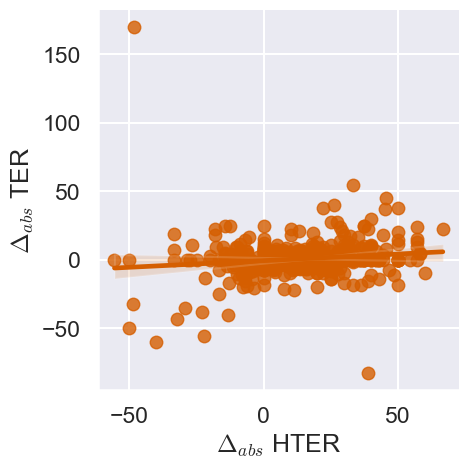}
            \caption{$\Delta_{abs}$ HTER and TER}
        \end{subfigure}
    \qquad\qquad\qquad
    \begin{subfigure}[t]{0.3\textwidth}
            \centering
            \includegraphics[width=\linewidth, 
            ]{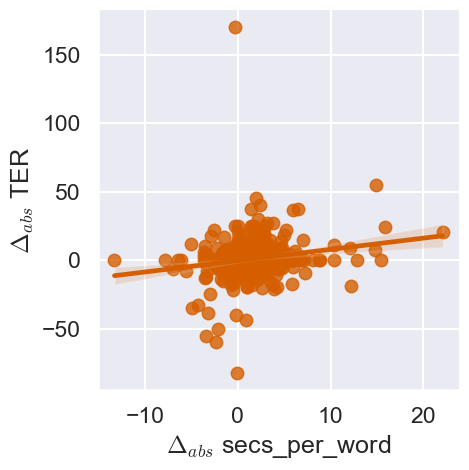}
            \caption{$\Delta_{abs}$ secs\_per\_word and TER}
        \end{subfigure}
    \caption{Scatter plots with overlaid regression lines on all datasets and languages for differences between feminine and masculine scores.}
    \label{fig:scatterplots_comet_ter}
\end{figure}

\subsubsection{BLEU Results per dataset}

We report language, users, and dataset-wise results of the correlations between the automatic metric BLEU and the human-centric metrics HTER and secs\_per\_word. Similar trends are also shown for COMET and TER, as discussed in Appendix \ref{subsubsec:correlations-ter-comer}.

Pearson correlation coefficients for each combination are shown in Table \ref{tab:correlation_r}. Language-wise correlations on \textsc{mtgen-a} are shown in Figure \ref{fig:scatterplots_mt_geneval_a} while dataset-wise correlations on \textsc{mtgeneval\_un} and \textsc{must-she} for \textit{en-it} are shown in Figure \ref{fig:scatterplots_it}.

\begin{table}[t]
\footnotesize
\centering
\setlength{\tabcolsep}{2.2pt}
\begin{tabular}{ll|c|c|c}
\toprule
&       \textbf{Pearson-$r$}       & \textsc{bleu-hter} & \textsc{bleu-spw} & \textsc{hter-spw} \\ 

\midrule
\textit{en-it} &  \textsc{mtgen-un} & 0.18   &  0.03$^\times$ & 0.50 \\
\textit{en-it} &  \textsc{must-she} & -0.14  & -0.22 & 0.48  \\
\cmidrule{2-5}
\textit{en-it} & \textsc{mtgen-a} (P)  & -0.22  &  -0.18 & 0.54 \\
\textit{en-it} & \textsc{mtgen-a} (S) & -0.24  &  -0.31 & 0.51 \\
\cmidrule{2-5}
\textit{en-es} & \textsc{mtgen-a}  & -0.44  &  -0.27 & 0.49 \\
\textit{en-de} & \textsc{mtgen-a}  &  0.19  &  0.03$^\times$  & 0.50 \\
\bottomrule
\end{tabular}
\setcounter{figure}{7}
\caption{Pearson R Coefficients of correlations between $\Delta_{abs}$ BLEU, $\Delta_{abs}$ HTER and $\Delta_{abs}$ SPW (secs\_per\_word), for the different datasets and languages analyzed in the paper. \underline{Non}-statistically significant results are indicated with $^\times$.
}
\label{tab:correlation_r}
\end{table}

In Section \ref{sec:discussion} , we elaborated on the weak correlations between automatic metrics such as BLEU scores and temporal and technical effort metrics such as HTER and seconds per word (SPW). When looking at the correlation results for each dataset, we observe similar trends: only HTER and SPW are moderately correlated while automatic and temporal/technical effort metrics exhibit no or weak correlation, with also some non-statically significant results. Therefore, the conclusions drawn when looking at aggregated statistics are similar to those obtained individually for each dataset.

\begin{figure*}
    \centering
    \captionsetup{labelformat=empty}
    \caption{\textbf{\textsc{mtgen-a} \textit{en-it} (P)}}
    \captionsetup{labelformat=parens}
    \begin{subfigure}[t]{0.325\textwidth}
            \centering
            \includegraphics[width=\linewidth, 
            ]{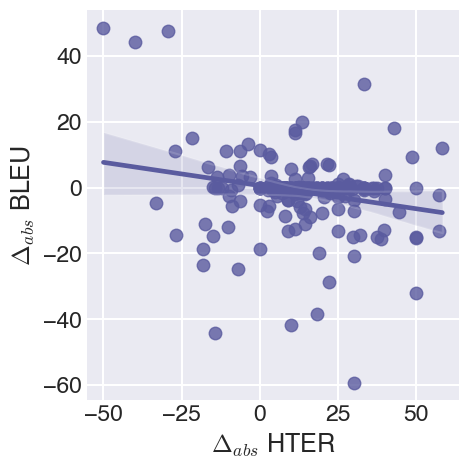}
            \caption{$\Delta_{abs}$ HTER and BLEU}
        \end{subfigure}
    \begin{subfigure}[t]{0.325\textwidth}
            \centering
            \includegraphics[width=\linewidth, 
            ]{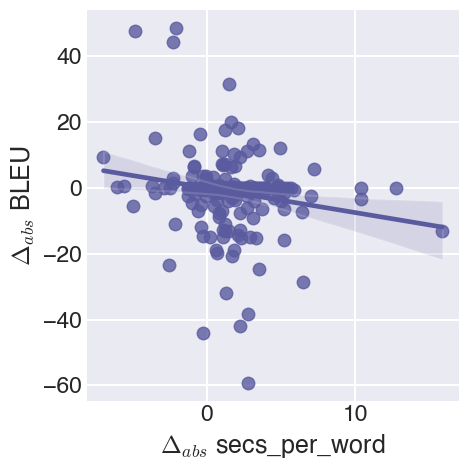}
            \caption{$\Delta_{abs}$ secs\_per\_word and  BLEU}
        \end{subfigure}
    \begin{subfigure}[t]{0.325\textwidth}
            \centering
            \includegraphics[width=\linewidth, 
            ]{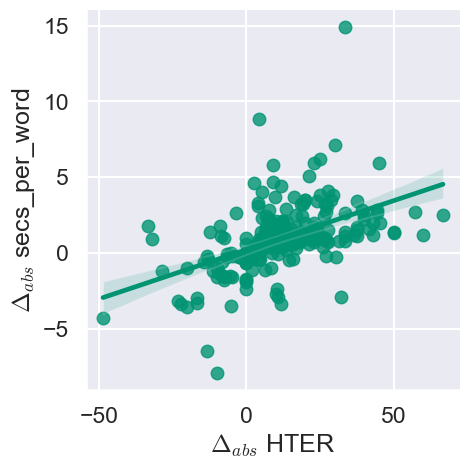}
            \caption{$\Delta_{abs}$ HTER and secs\_per\_word}
        \end{subfigure}\\
    \captionsetup{labelformat=empty}
    \caption{\textbf{\textsc{mtgen-a} \textit{en-es}}}
    \captionsetup{labelformat=parens}
    \begin{subfigure}[t]{0.325\textwidth}
            \centering
            \includegraphics[width=\linewidth, 
            ]{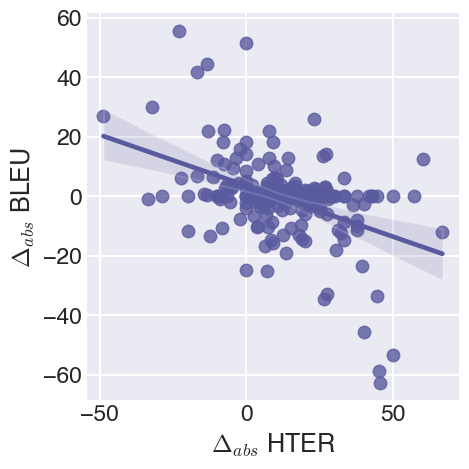}
            \caption{$\Delta_{abs}$ HTER and BLEU}
        \end{subfigure}
    \begin{subfigure}[t]{0.325\textwidth}
            \centering
            \includegraphics[width=\linewidth, 
            ]{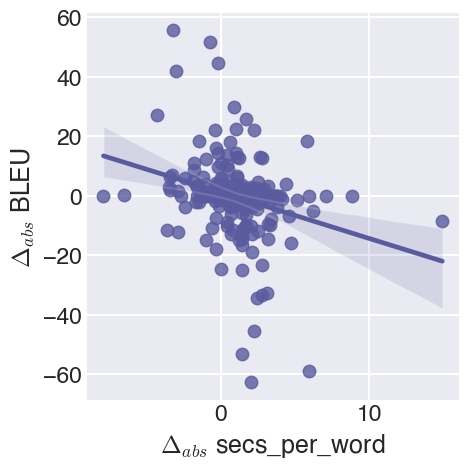}
            \caption{$\Delta_{abs}$ secs\_per\_word and  BLEU}
        \end{subfigure}
    \begin{subfigure}[t]{0.325\textwidth}
            \centering
            \includegraphics[width=\linewidth, 
            ]{images/appendix/correlations/mtgeneval_a_es_hter_te.png}
            \caption{$\Delta_{abs}$ HTER and secs\_per\_word}
        \end{subfigure}\\
    \captionsetup{labelformat=empty}
    \caption{\textbf{\textsc{mtgen-a} \textit{en-de}}}
    \captionsetup{labelformat=parens}
    \begin{subfigure}[t]{0.325\textwidth}
            \centering
            \includegraphics[width=\linewidth, 
            ]{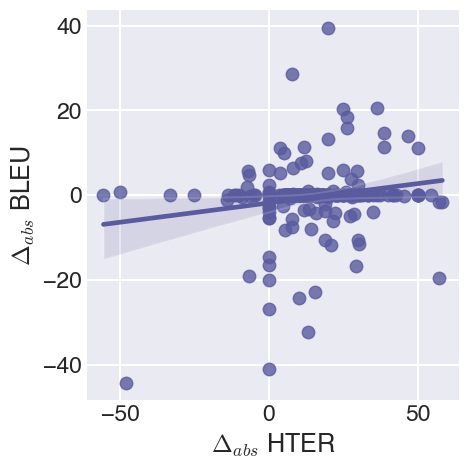}
            \caption{$\Delta_{abs}$ HTER and BLEU}
        \end{subfigure}
    \begin{subfigure}[t]{0.325\textwidth}
            \centering
            \includegraphics[width=\linewidth, 
            ]{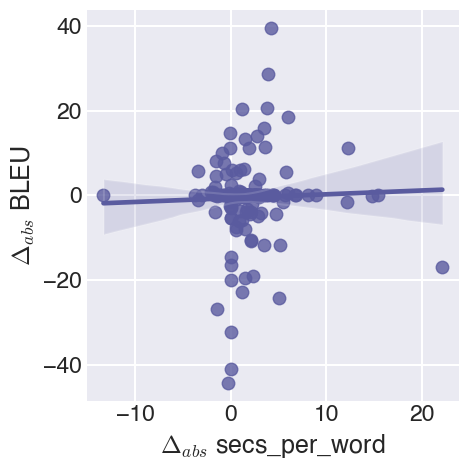}
            \caption{$\Delta_{abs}$ secs\_per\_word and  BLEU}
        \end{subfigure}
    \begin{subfigure}[t]{0.325\textwidth}
            \centering
            \includegraphics[width=\linewidth, 
            ]{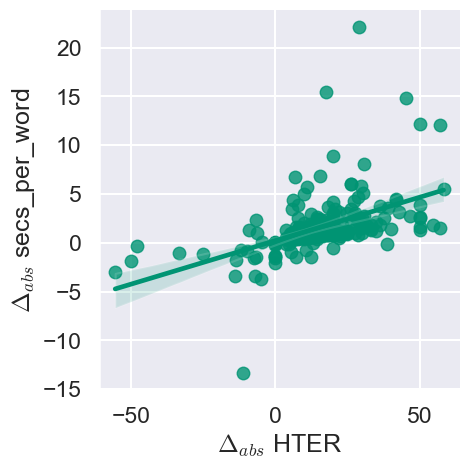}
            \caption{$\Delta_{abs}$ HTER and secs\_per\_word}
        \end{subfigure}
    \setcounter{figure}{6}
    \caption{Scatter plots with overlaid regression lines for all languages on \textsc{mtgen-a}.}
    \label{fig:scatterplots_mt_geneval_a}
\end{figure*}

\begin{figure*}
    \centering
    \captionsetup{labelformat=empty}
    \caption{\textbf{\textsc{mtgen-a} \textit{en-it} (S)}}
    \captionsetup{labelformat=parens}
    \begin{subfigure}[t]{0.325\textwidth}
            \centering
            \includegraphics[width=\linewidth, 
            ]{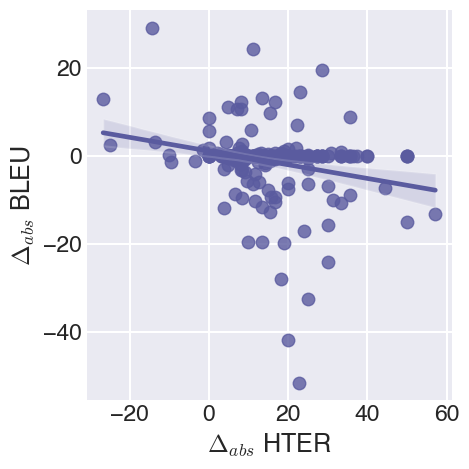}
            \caption{$\Delta_{abs}$ HTER and BLEU}
        \end{subfigure}
    \begin{subfigure}[t]{0.325\textwidth}
            \centering
            \includegraphics[width=\linewidth, 
            ]{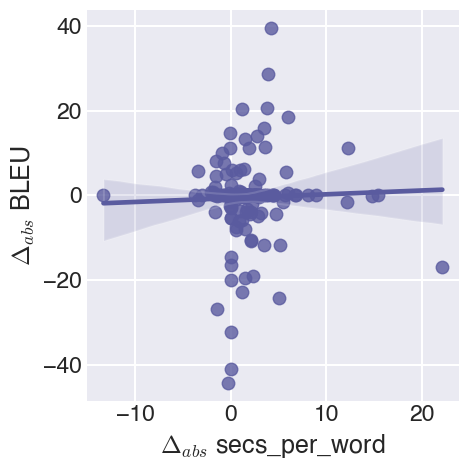}
            \caption{$\Delta_{abs}$ secs\_per\_word and  BLEU}
        \end{subfigure}
    \begin{subfigure}[t]{0.325\textwidth}
            \centering
            \includegraphics[width=\linewidth, 
            ]{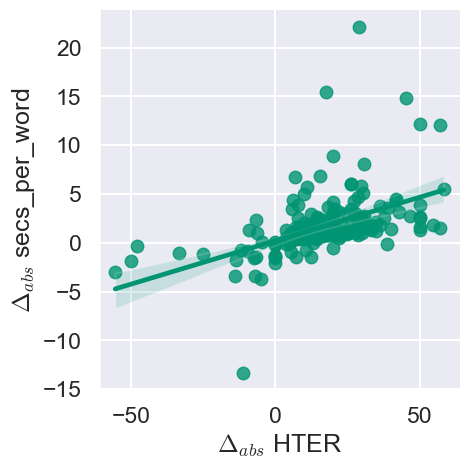}
            \caption{$\Delta_{abs}$ HTER and secs\_per\_word}
        \end{subfigure}\\
    \captionsetup{labelformat=empty}
    \caption{\textbf{\textsc{mtgen-un}}}
    \captionsetup{labelformat=parens}
    \begin{subfigure}[t]{0.325\textwidth}
            \centering
            \includegraphics[width=\linewidth, 
            ]{images/appendix/correlations/mtgeneval_a_s_hter_bleu.png}
            \caption{$\Delta_{abs}$ HTER and BLEU}
        \end{subfigure}
    \begin{subfigure}[t]{0.325\textwidth}
            \centering
            \includegraphics[width=\linewidth, 
            ]{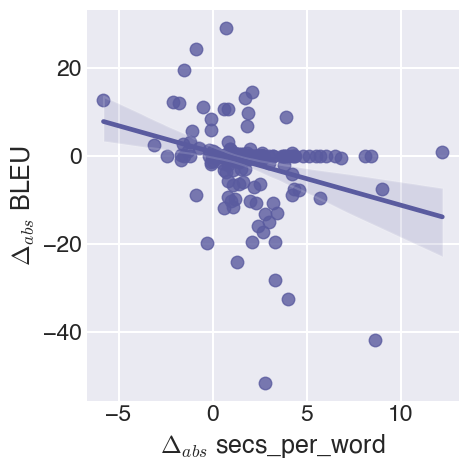}
            \caption{$\Delta_{abs}$ secs\_per\_word and  BLEU}
        \end{subfigure}
    \begin{subfigure}[t]{0.325\textwidth}
            \centering
            \includegraphics[width=\linewidth, 
            ]{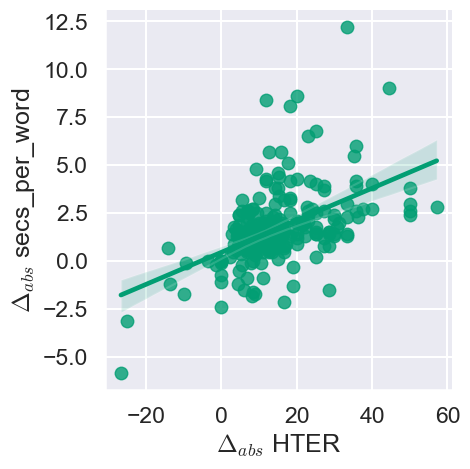}
            \caption{$\Delta_{abs}$ HTER and secs\_per\_word}
        \end{subfigure}\\
    \captionsetup{labelformat=empty}
    \caption{\textbf{\textsc{must-she}}}
    \captionsetup{labelformat=parens}
    \begin{subfigure}[t]{0.325\textwidth}
            \centering
            \includegraphics[width=\linewidth, 
            ]{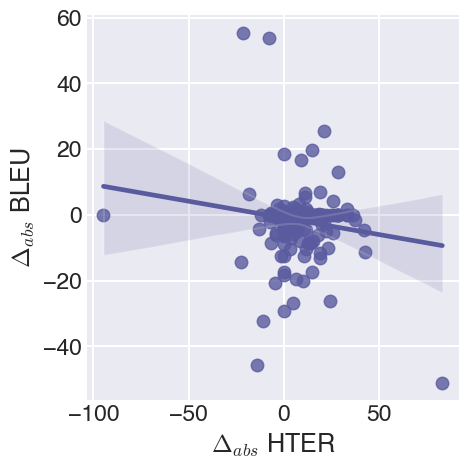}
            \caption{$\Delta_{abs}$ HTER and BLEU}
        \end{subfigure}
    \begin{subfigure}[t]{0.325\textwidth}
            \centering
            \includegraphics[width=\linewidth, 
            ]{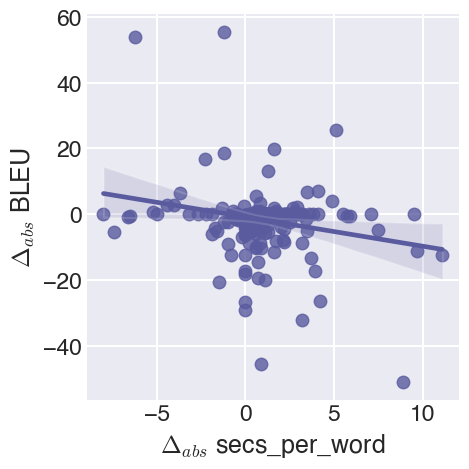}
            \caption{$\Delta_{abs}$ secs\_per\_word and  BLEU}
        \end{subfigure}
    \begin{subfigure}[t]{0.325\textwidth}
            \centering
            \includegraphics[width=\linewidth, 
            ]{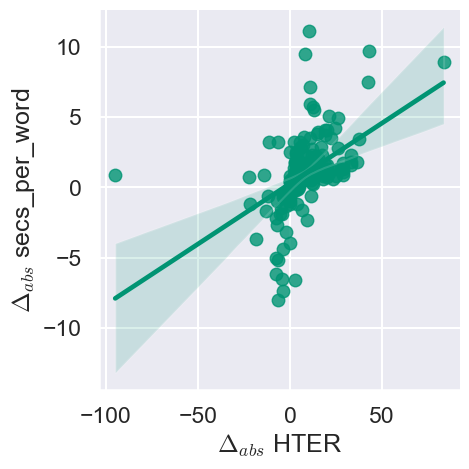}
            \caption{$\Delta_{abs}$ HTER and secs\_per\_word}
        \end{subfigure}
    \setcounter{figure}{7}
    \caption{Scatter plots with overlaid regression lines on \textsc{mtgen-un} (S), \textsc{mtgen-un} and \textsc{must-she} for \textit{en-it}.}
    \label{fig:scatterplots_it}
\end{figure*}

\end{document}